\documentclass[a4paper,11pt]{article}
\usepackage[utf8]{inputenc}
\usepackage[english]{babel} 
\usepackage[a4paper,top=25mm,bottom=30mm,left=25mm,right=25mm]{geometry}

\usepackage{lineno,hyperref}
\usepackage{amsmath}
\usepackage{amssymb}
\usepackage{lscape}
\usepackage{microtype}
\usepackage{array}
\usepackage{multirow}
\usepackage{graphicx}
\usepackage[caption=false,font=scriptsize]{subfig}
\usepackage[table,xcdraw,dvipsnames]{xcolor}
\usepackage{float}
\usepackage{longtable}
\modulolinenumbers[5]

\graphicspath{{./figures/}}
\DeclareGraphicsExtensions{.pdf,.jpeg,.png,.jpg}

\newcommand{\norm}[1]{\left\lVert#1\right\rVert}
\newcommand{\abs}[1]{\left\lvert#1\right\rvert}
\newcommand{\hmax}{H_{\mathrm{max}}}
\newcommand{\rhof}[1]{(\rho_{\partial F})_{\mathrm{#1}}}
\newcommand{\rhops}{\rho_{\partial S_o}}
\newcommand{\com}[1]{\mathcal{F}_{\mathrm{#1}}}

\newcommand{\basin}[1]{\mathcal{B}_{\mathrm{#1}}}
\newcommand{\minset}[1]{\mathcal{M}_{\mathrm{#1}}}
\newcommand{\basinf}[1]{(\mathcal{B}_\mathrm{F})_{\mathrm{#1}}}
\newcommand{\R}[1]{\mathbb{R}^{#1}}

\makeatletter
\renewcommand\paragraph{\@startsection{paragraph}{4}{\z@}%
            {-2.5ex\@plus -1ex \@minus -.25ex}%
            {1.25ex \@plus .25ex}%
            {\normalfont\normalsize\bfseries}}
\makeatother
\date{\today}
\title{  
  \vfill
  \textbf{Characterization of Constrained Continuous Multiobjective Optimization Problems:\\ A Feature Space Perspective} \\[1cm]
}
\author{Aljoša Vodopija, Tea Tušar, Bogdan Filipič\\[10pt]
Jožef Stefan Institute \emph{and}\\
Jožef Stefan International Postgraduate School\\
Ljubljana, Slovenija \\[10pt]
{\em aljosa.vodopija@ijs.si, tea.tusar@ijs.si}, \\ 
{\em bogdan.filipic@ijs.si}}

\begin{document}

\maketitle
\vfill
\thispagestyle{empty}

\begin{abstract}
Despite the increasing interest in constrained multiobjective optimization in recent years, constrained multiobjective optimization problems (CMOPs) are still insufficiently understood and characterized. For this reason, the selection of appropriate CMOPs for benchmarking is difficult and lacks a formal background. We address this issue by extending landscape analysis to constrained multiobjective optimization. By employing four exploratory landscape analysis techniques, we propose 29 landscape features (of which 19 are novel) to characterize CMOPs. These landscape features are then used to compare eight frequently used artificial test suites against a recently proposed suite consisting of real-world problems based on physical models. The experimental results reveal that the artificial test problems fail to adequately represent some realistic characteristics, such as strong negative correlation between the objectives and the overall constraint violation. Moreover, our findings show that all the studied artificial test suites have advantages and limitations, and that no ``perfect'' suite exists. Benchmark designers can use the obtained results to select or generate appropriate CMOP instances based on the characteristics they want to explore.
\end{abstract}
\vspace{1cm}

\newpage
\thispagestyle{empty}

\newpage
\pagestyle{plain}

\section{Introduction} 
\label{sec:introduction}

Many real-world continuous optimization problems involve multiple conflicting objectives that need to be optimized simultaneously and constraints that need to be respected~\cite{Ma2019}. Such optimization problems are called \emph{constrained multiobjective optimization problems} (CMOPs) and have recently gained much interest in the optimization community. Indeed, many novel approaches for constraint handling and new benchmark suites of CMOPs have been proposed in the last three years~\cite{Filipic21}.

For this reason, never before has there been such a strong need for problem analysis that would inform the selection of CMOPs for benchmarking multiobjective optimization algorithms (MOA). Currently, the benchmark suites are still insufficiently understood, and their characteristics remain unrevealed. In addition to characteristics describing the number of variables, objectives and constraints, type of objective and constraint functions, and geometric properties of Pareto front shapes, there are only few and limited techniques proposed to explore the search space of CMOPs, especially from the field of exploratory landscape analysis (ELA)~\cite{Picard2021, Kerschke2016, Malan2015}.

There are only limited studies on ELA in the multiobjective combinatorial context~\cite{Verel2013, Daolio2017, Liefooghe2019}, an initial study in continuous multiobjective optimization~\cite{Liefooghe2021}, and some first attempts to visualize bi-objective continuous problem landscapes with up to three variables~\cite{Fonseca1995, Kerschke2016, Kerschke2017, Schapermeier2020}. The situation is even worse in constrained optimization, with a single preliminary study on the characterization of constrained continuous single-objective optimization problems~\cite{Malan2015} and an attempt to extend this work to multiobjective optimization~\cite{Picard2021}. The latter work studied five ELA features, including the feasibility ratio, disjointedness of feasible regions, and relationships between the objectives and constraints. Although the study considers ELA techniques for multiple objectives and constraints simultaneously, many other crucial aspects still need to be considered. For example, recently proposed constraint handling techniques assume that CMOP violation landscapes consist of several ``suboptimal'' subregions---what we will call \emph{violation multimodality}---that severely aggravate algorithm performance~\cite{Zhu2020}. While the multimodality for unconstrained multiobjective optimization has been extensively studied in~\cite{Grimme21}, the violation multimodality has not yet been studied from the ELA perspective~\cite{Filipic21}.

Due to the lack of appropriate techniques to characterize CMOPs, preparing a sound and well-designed experimental setup for constrained multiobjective optimization is a challenging task. A poorly designed benchmark might lead to inadequate conclusions about CMOP landscapes and MOA performance. Two steps are required to resolve this issue: 1) providing new ELA techniques specialized for CMOPs and 2) using them to reveal the relationship between CMOP characteristics and MOA performance~\cite{Munoz2015a}. In this paper, we address step 1) by extending various ELA techniques to constrained multiobjective optimization, and provide a thorough analysis of the available test suites of CMOPs. In contrast, step 2) is currently under investigation and not a part of this paper due to limited space.

The contributions of this work are as follows:
\begin{itemize}
    \item We generalize the concept of a fitness landscape to CMOPs and introduce two mathematical definitions to formally describe violation multimodality in constrained multiobjective optimization.
    \item We use various ELA techniques including space-filling design, information content, random and adaptive walks to derive 29 ELA features, of which 19 are novel. These features are then used to characterize violation multimodality and smoothness and measure the conflicting nature between the objectives and constraints. In contrast to the related work, we also discuss the sensitivity and scalability of the proposed techniques.
    \item We compare the most frequently used artificial test suites against selected real-world problems from the RCM suite---a novel test suite consisting of real-world CMOPs based on physical models~\cite{Kumar2021}. Specifically, we assess whether the studied artificial test problems comprehensively represent the characteristics observed in the RCM problems. Note that only continuous and low-dimensional RCM problems are used in this study. The test suites are also assessed with respect to the representatives of various CMOP characteristics by contrasting them against the set of all the considered CMOPs.
\end{itemize}
Because of space limitations, we present only selected results in this paper. The interested reader can find the complete results online\footnote{https://vodopijaaljosa.github.io/cmop-web/}.

The rest of this paper is organized as follows. In Section~\ref{sec:theoretical_background}, we provide the theoretical background for landscape analysis of CMOPs, while the theoretical advances are introduced in Section~\ref{sec:theoretical_advances}. Then, in Section~\ref{sec:methodology}, we introduce the ELA features for CMOPs and discuss the methodology used to derive them. Section~\ref{sec:experiments} provides details on the experimental setup, presents the results, evaluates the existing test suites of CMOPs, and discusses the sensitivity and scalability of the applied methodology. Finally, Section~\ref{sec:conclusions} concludes the work with a summary of findings and ideas for future work.

\section{Theoretical background} 
\label{sec:theoretical_background}

This section provides the theoretical background for this study. After the definitions of the constrained multiobjective optimization problem and constraint violation, we present the concept of the fitness and violation landscape. Finally, we discuss the topological property of connectedness.

\subsection{Constrained multiobjective optimization problems}

A CMOP is, without loss of generality, formulated as:
\begin{equation} \label{eq:cmop}
\begin{split}
    \text{minimize} \quad &f_m(x), \quad m = 1, \dots, M \\
    \text{subject to} \quad &g_i(x) \leq 0, \quad i = 1, \dots, I\\
    \quad &h_i(x) = 0, \quad i = I + 1, \dots, I + J\\
\end{split}
\end{equation}
where $x = (x_1, \dots, x_D)$ is a \emph{search vector}, $f_m: S \rightarrow \R{}$ are \emph{objective functions}, $g_i: S \rightarrow \R{}$ \emph{inequality constraint functions}, $h_i: S \rightarrow \R{}$ \emph{equality constraint functions}, $S \subseteq \R{D}$ is a \emph{search space} of dimension $D$, and $M$, $I$ and $J$ are numbers of objectives, inequality and equality constraints, respectively. 

In continuous optimization, the equality constraints are usually reformulated into inequality constraints as follows:
\begin{equation}
    g_i(x) = \lvert h_i(x) \rvert - \eta \leq 0, \quad i = I + 1, \dots,  I + J
\end{equation}
where $\eta > 0$ is a user-defined tolerance value to relax equality constraints (in our study it is set to $10^{-4}$). Using this definition, all constraints can be treated as inequality constraints. A solution $x$ is said to be \emph{feasible} if it satisfies all the constraints $g_i(x) \leq 0$ for $i = 1, \dots, I + J$.

One of the most important concepts in constrained optimization is the notion of the \emph{constraint violation}. For a single constraint $g_i$ it is defined as $v_i(x) = \max(0, g_i(x))$. For all constraints together it is combined as
\begin{equation}
    v(x) = \sum_i^{I+J} v_i(x)
\end{equation}
into the \emph{overall constraint violation}. A solution $x$ is feasible iff its overall constraint violation equals zero ($v(x) = 0$). Note that other definitions for overall constraint violation exist, and their use would impact the analysis performed in this study. However, the proposed definition for the overall constraint violation is by far the most commonly adopted in constrained optimization~\cite{Filipic21}, and as such, it represents the most appropriate choice.

A feasible solution $x \in S$ \emph{dominates} another solution $y \in S$ iff $f_m(x) \leq f_m(y)$ for all $1 \leq m \leq M$ and $f_m(x) < f_m(y)$ for at least one $1 \leq m \leq M$. Additionally, a solution $x^* \in S$ is \emph{Pareto optimal} if there is no solution $x \in S$ such that it dominates $x^*$.

The set of all feasible solutions is called the \emph{feasible region} and is denoted by $F = \{x \in S \mid v(x) = 0\}$. All nondominated feasible solutions represent a \emph{Pareto-optimal set}, $S_\text{o}$. The image of the Pareto-optimal set in the objective space is the \emph{Pareto front} and is denoted here by $P_\text{o} = \{f(x) \mid x \in S_\text{o}\}$.

\subsection{Fitness and violation landscapes}
\label{sec:fitenss_and_violation_landscapes}

The concept of a \emph{fitness landscape} was formally defined in~\cite{Stadler2002} by identifying the following three key elements:
\begin{itemize}
    \item $S \subseteq \R{D}$ is the search space,
    \item $f: S \rightarrow \R{}$ is the objective function,
    \item $d: S \times S \rightarrow \R{}$ is the distance metric.
\end{itemize}

The distance metric $d$ is used to quantify the similarity between solutions in the search space and to define \emph{neighborhoods}. In continuous optimization, the Euclidean distance is often used. Close solutions are those whose Euclidean distance is within a predefined threshold, $\delta$. Formally, the solution $y \in S$ is part of the neighborhood of the solution $x \in S$, denoted by $\mathcal{N}(x, \delta)$, iff $d(x, y) \leq \delta$. 

In~\cite{Malan2015}, an adaptation of the fitness landscape was proposed for constrained optimization problems by replacing the objective function with the overall constraint violation function. The resulting idea is known as the \emph{violation landscape}.

\subsection{Connectedness}

A set $X \in S$ of a topological space is said to be \emph{disconnected} if it can be represented as the union of two or more disjoint non-empty open subsets from $X$. Otherwise, $X$ is said to be \emph{connected}.

The maximal connected subsets (with respect to the inclusion order) of a non-empty space are called \emph{connected components}. The components of any topological space $X$ form a partition of $X$---they are disjoint, non-empty, and their union is the entire space.

\section{Theoretical advances}
\label{sec:theoretical_advances}

In this section, we overview theoretical advances in landscape analysis for CMOPs. After defining constrained multiobjective problem landscapes, we introduce feasible components and basins of attraction in the violation landscape. Finally, we conclude with three examples depicting these novel concepts.

\subsection{Constrained multiobjective problem landscapes} 
\label{sec:constrained_multiobjective_problem landscapes}

Following the definitions of fitness and violation landscapes from Section~\ref{sec:fitenss_and_violation_landscapes}, we generalize these concepts to CMOPs by including multiple objective functions and the overall constraint violation function simultaneously. A \emph{constrained multiobjective problem landscape} is defined with the following four elements:
\begin{itemize}
    \item $S \subseteq \R{D}$ is the search space,
    \item $f: S \rightarrow \R{M}$ is the objective vector function,
    \item $v: S \rightarrow \R{}$ is the overall constraint violation function,
    \item $d: S \times S \rightarrow \R{}$ is the distance metric.
\end{itemize}

The notion of a neighborhood can be applied to CMOP landscapes without any modification. For example, a solution $x^* \in S$ is a \emph{local Pareto-optimal solution} if it is feasible and there exists a $\delta > 0$ such that no feasible solution from the neighborhood $\mathcal{N}(x^*, \delta)$ dominates $x^*$. The set of all local Pareto-optimal solutions will be denoted by $S_\text{l}$. It is obvious that $S_\text{o} \subseteq S_\text{l}$.

In~\cite{Zhu2020}, two topological structures concerning violation multimodality were identified that might cause premature algorithm stagnation: feasible subregions and areas that locally violate the constraints the least. The feasible region can consist of multiple disconnected subregions. The more such subregions there are, the harder it is for the algorithm to find a good approximation for the whole Pareto-optimal set. Moreover, the search space of a CMOP can have many areas that locally violate the constraints the least and are infeasible. Again, many such areas might aggravate the performance of an algorithm since it can get stuck in the infeasible region. Although these concepts are essential, they are not formally defined, which can lead to ambiguous interpretations. To overcome this situation, we provide a rigorous mathematical formulation and, in this way, facilitate their further consideration.

\subsection{Feasible components} 
\label{sec:feasible_components}

A connected component $\com{} \subseteq F$ of the feasible region is called a \emph{feasible component}. Note that all feasible components form a partition of the feasible region. With feasible components, we provide a formal definition of feasible subregions. A feasible component can be treated as ``suboptimal'' when it contains no Pareto-optimal solutions. Obviously, if a problem landscape contains several ``suboptimal'' feasible components, it might be harder for an optimizer to find good approximations for all Pareto-optimal solutions.

\subsection{Basins of attraction in the violation landscape} 
\label{sec:basins_attraction}

Basins of attraction in the fitness landscape are well-known objects widely used to characterize unconstrained single-objective optimization problems. In this paper, we reuse this concept in the violation landscape. However, we need to introduce some crucial notions first.

Similarly to the local Pareto-optimal solution, we can define a local ``optimal'' solution in the violation landscape---a solution that locally violates the constraints the least. Formally, $x^*$ is a \emph{local minimum-violation solution} if there exist a $\delta > 0$ such that $v(x^*) \leq v(x)$ for all $x \in \mathcal{N}(x^*,\delta)$. Additionally, if there exists no solution $x \in S$ such that $v(x^*) > v(x)$, then $x^*$ is a \emph{(global) minimum-violation solution}. The set of all local minimum-violation solutions will be denoted by $F_\text{l}$ and of all minimum-violation solutions by $F_\text{m}$. It is obvious that $F_\text{m} \subseteq F_\text{l}$ and if there exist feasible solutions in $S$, then $F_\text{m} = F$.

A connected component $\minset{} \subseteq F_\text{l}$ is called a \emph{local minimum-violation component} and it provides a formal definition for areas that locally violate the constraints the least. All the local minimum-violation components form a partition of $F_\text{l}$.

Now, we can finally define a basin of attraction. Let us consider an abstract local search procedure as a mapping from the search space to the set of local minimum-violation solutions, $\mu : S \rightarrow F_\text{l}$, such that $\mu(x) = x$ for all $x \in F_\text{l}$. Then, a \emph{basin of attraction} of a local minimum-violation component $\minset{}$ and local search $\mu$ is a subset of $S$ in which $\mu$ converges towards a solution from $\minset{}$, i.e., $\basin{}(\minset{}) = \{x \in S \mid \mu(x) \in \minset{}\}$. In this case, the local minimum-violation component $\minset{}$ is said to be an \emph{attractor} of the basin $\basin{}(\minset{})$. If there is only one basin in $S$, then the corresponding violation landscape is said to be \emph{unimodal}. Otherwise, it is \emph{multimodal}.

Each feasible component is a subset of one of the local minimum-violation components. The reverse is, in general, not true. However, if a local minimum-violation component contains feasible solutions, then it contains at least one feasible component as well. Consequently, one can discover feasible components by first identifying local minimum-violation components that contain feasible solutions.  

\subsection{Examples}

\begin{figure*}[!t]
    \centering
    
    \subfloat[C2-DTLZ2:\\ Dominance ratio]{\includegraphics[clip, trim={0, 10pt, 0, 25pt}, width=0.33\textwidth]{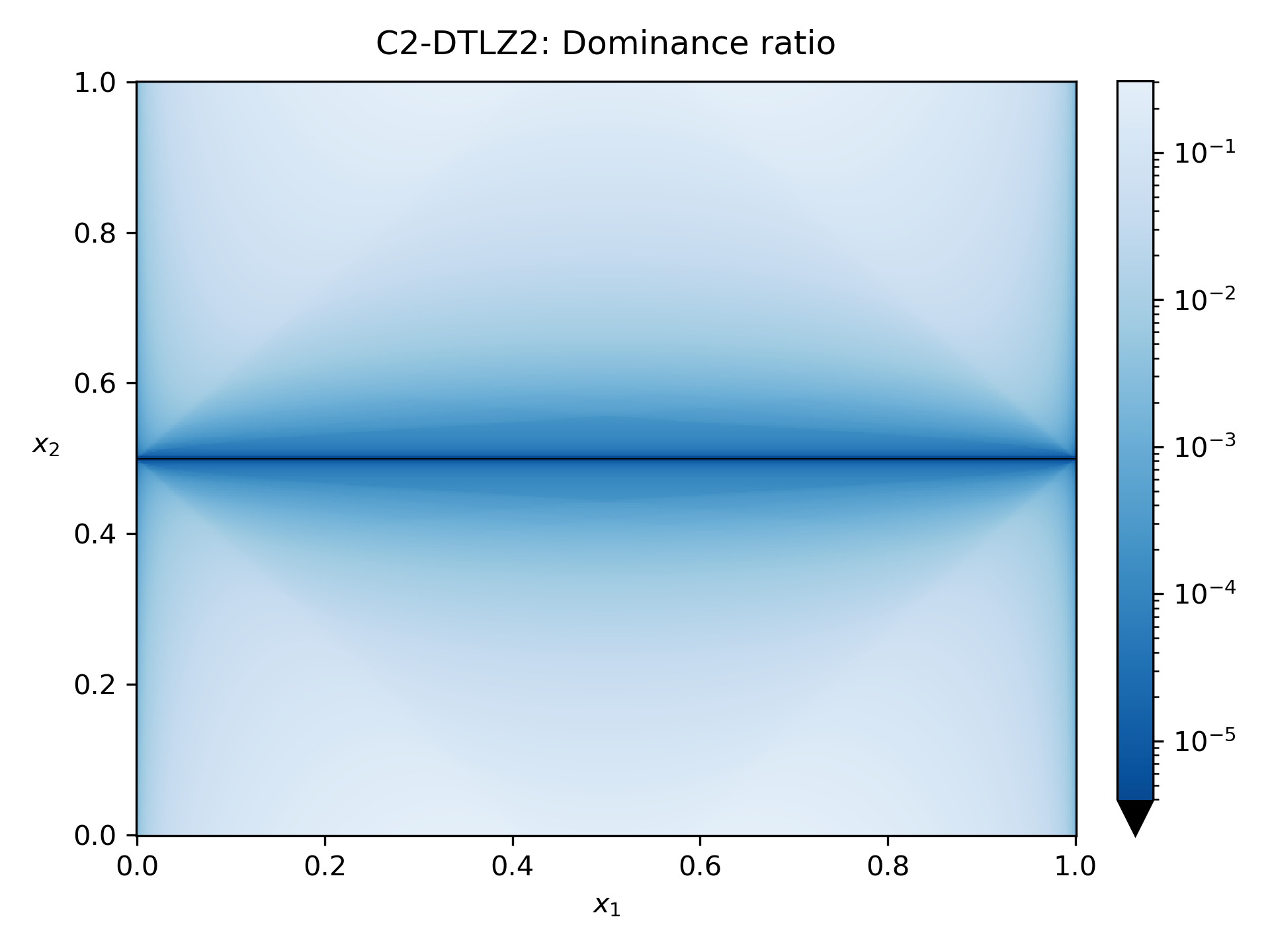}%
    \label{fig:c2dtlz2_fl}}
    \hfil
    \subfloat[C2-DTLZ2:\\ Violation landscape]{\includegraphics[clip, trim={0, 10pt, 0, 25pt}, width=0.33\textwidth]{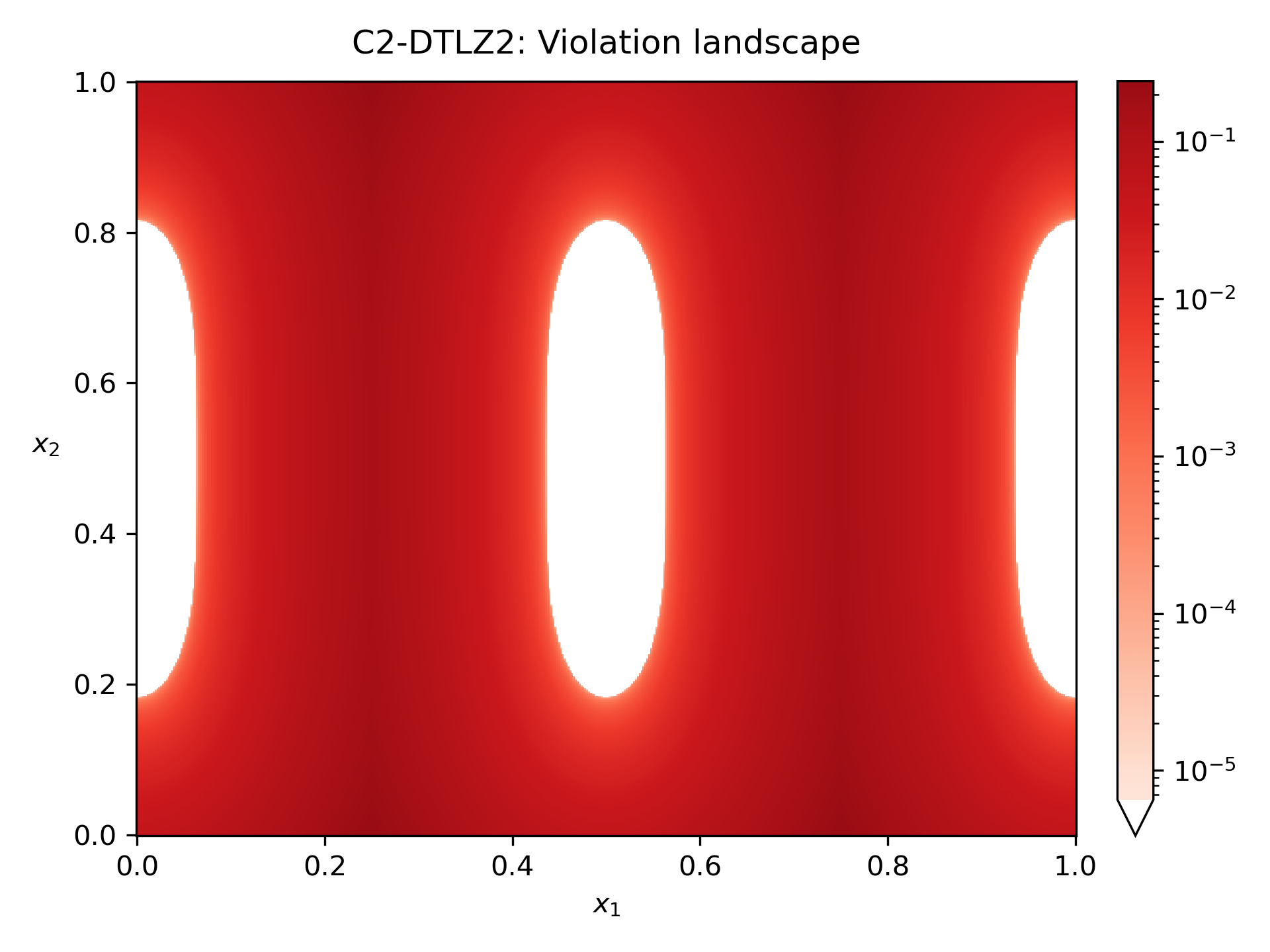}%
    \label{fig:c2dtlz2_vl}}
    \hfil
    \subfloat[C2-DTLZ2:\\ Problem landscape]{\includegraphics[clip, trim={0, 10pt, 0, 25pt}, width=0.33\textwidth]{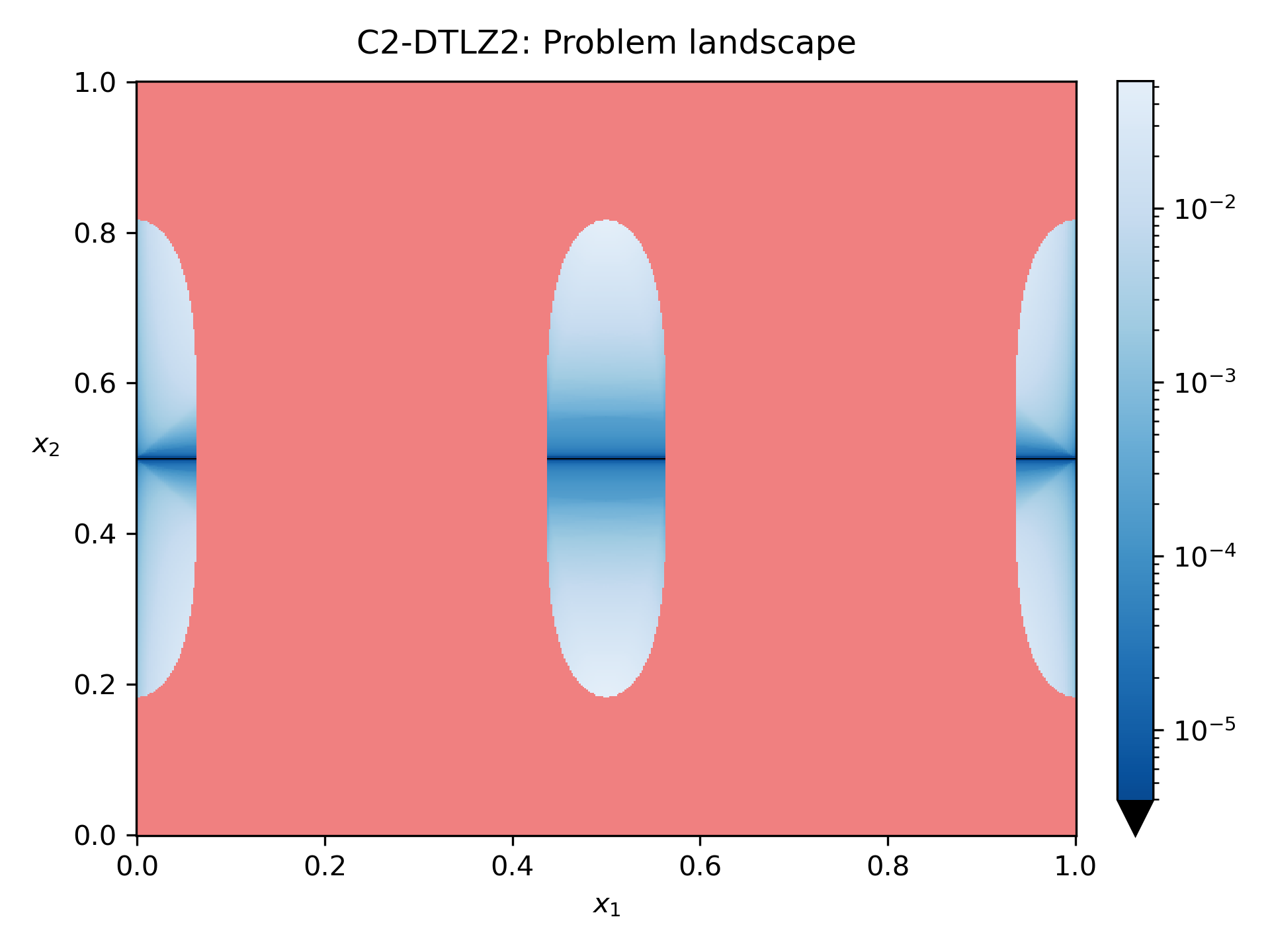}%
    \label{fig:c2dtlz2_pl}}

    \subfloat[MW6:\\ Dominance ratio]{\includegraphics[clip, trim={0, 10pt, 0, 25pt}, width=0.33\textwidth]{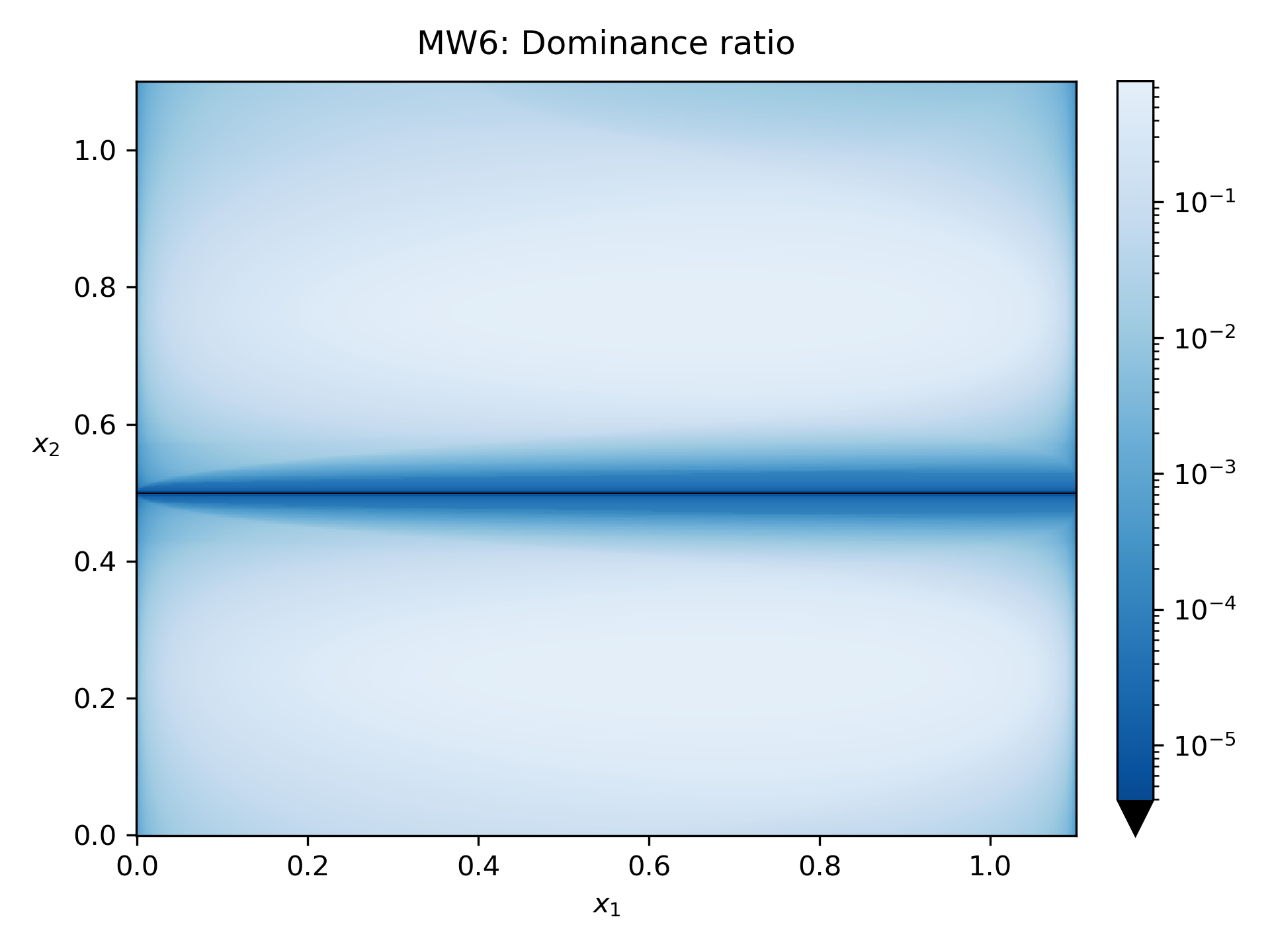}%
    \label{fig:mw6_fl}}
    \hfil
    \subfloat[MW6:\\ Violation landscape]{\includegraphics[clip, trim={0, 10pt, 0, 25pt}, width=0.33\textwidth]{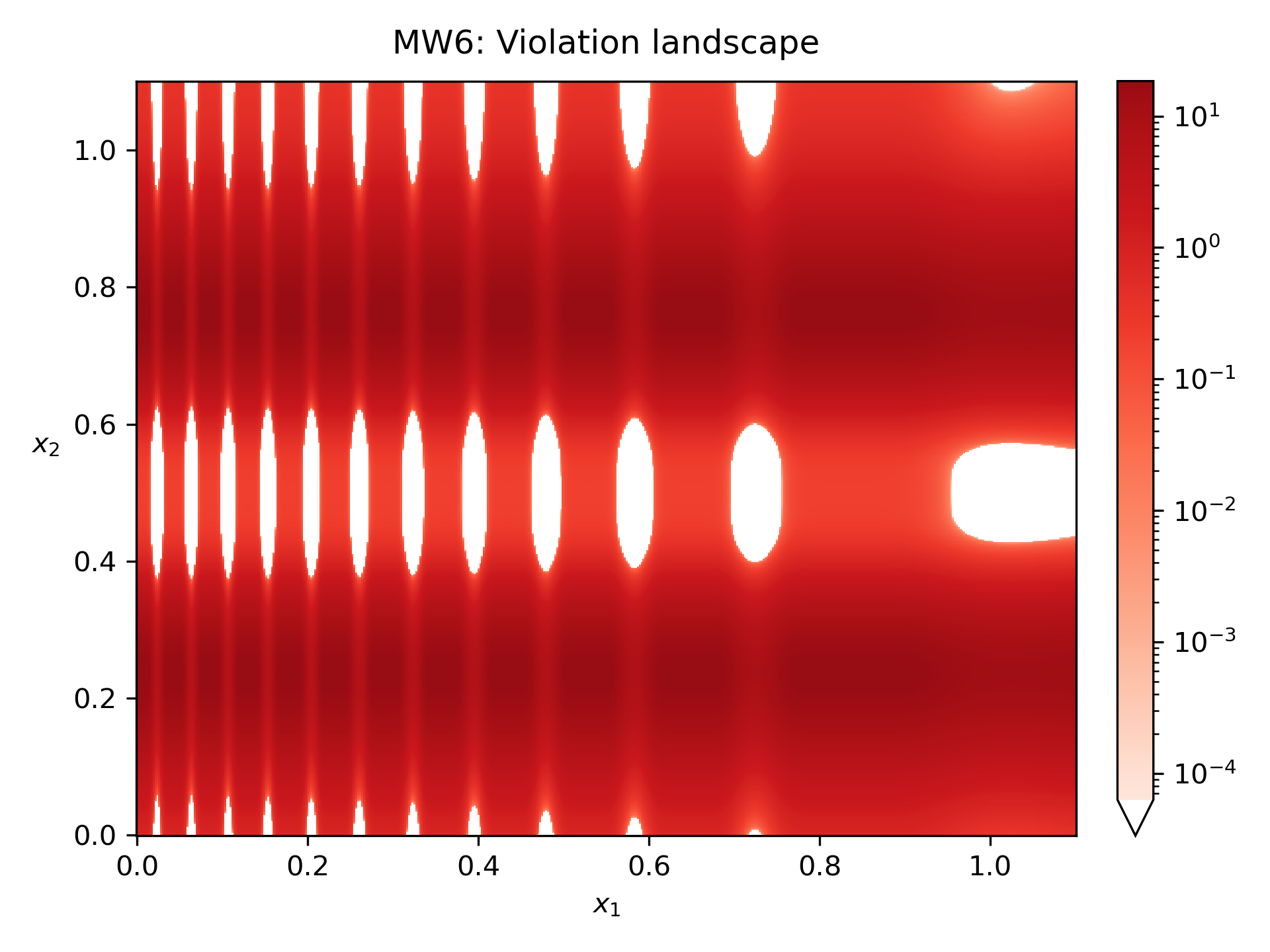}%
    \label{fig:mw6_vl}}
    \hfil
    \subfloat[MW6:\\ Problem landscape]{\includegraphics[clip, trim={0, 10pt, 0, 25pt}, width=0.33\textwidth]{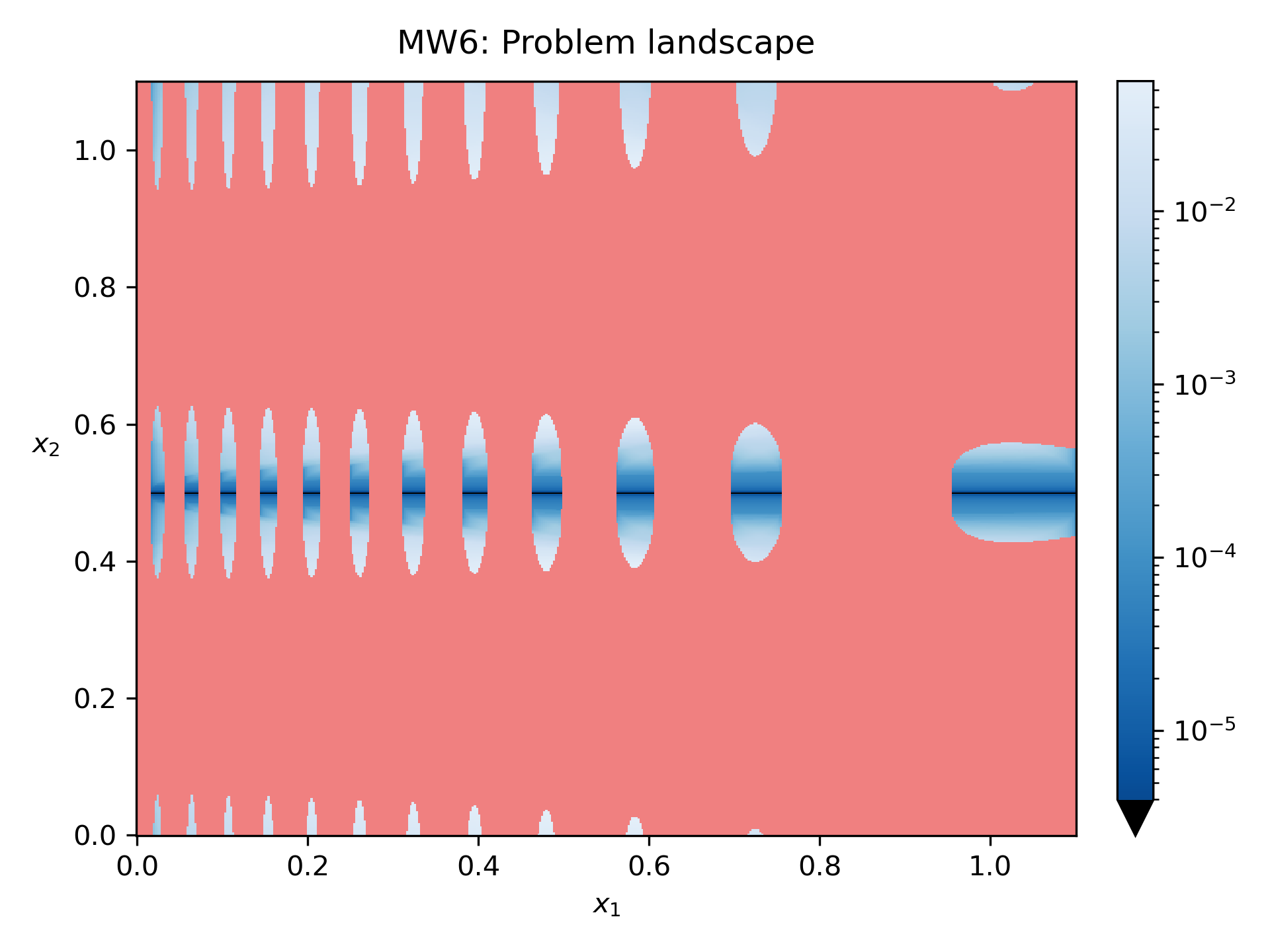}%
    \label{fig:mw6_pl}}
    
    \subfloat[DAS-CMOP1:\\ Dominance ratio]{\includegraphics[clip, trim={0, 10pt, 0, 25pt}, width=0.33\textwidth]{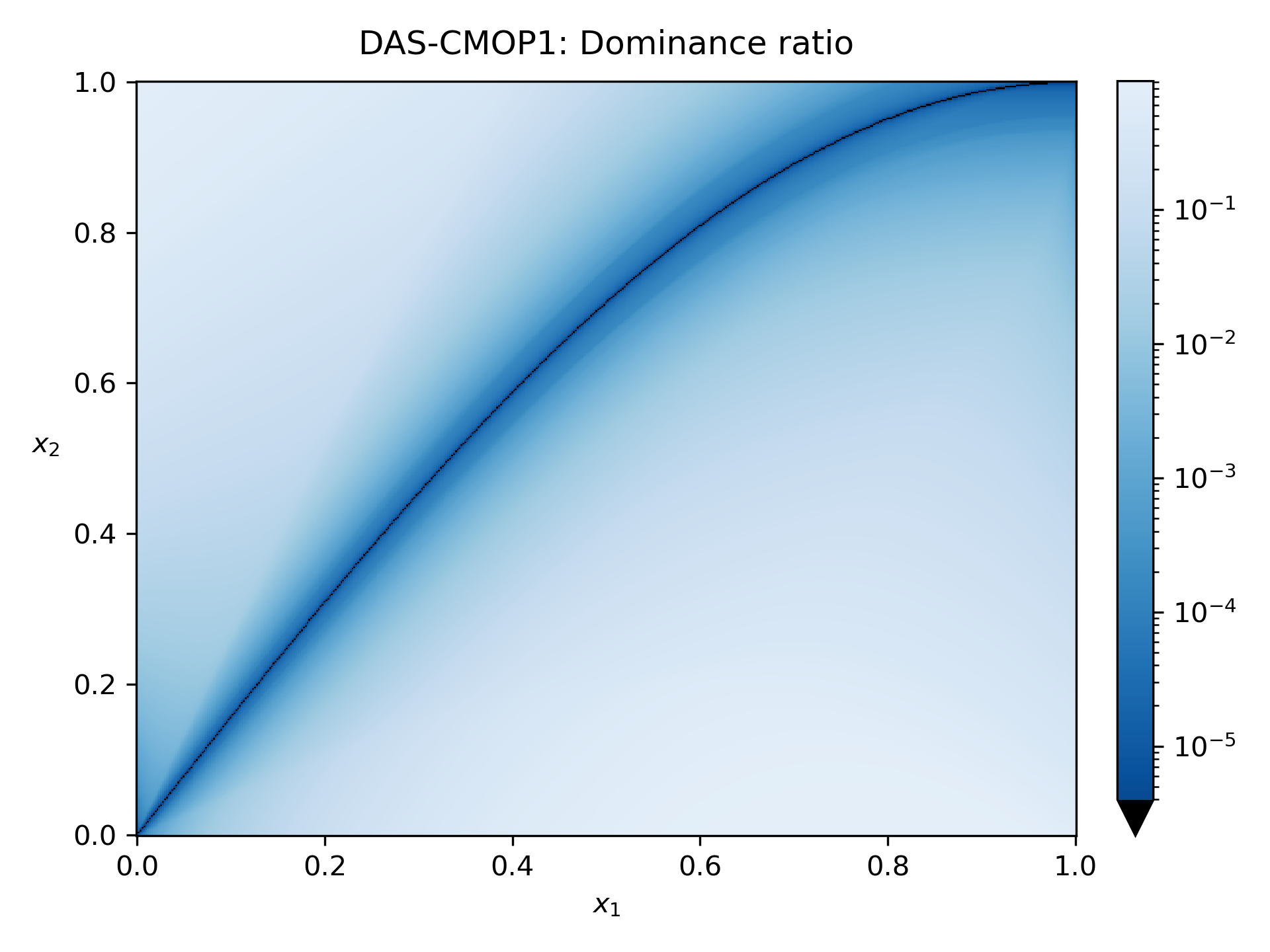}%
    \label{fig:dascmop1_fl}}
    \hfil
    \subfloat[DAS-CMOP1:\\ Violation landscape]{\includegraphics[clip, trim={0, 10pt, 0, 25pt}, width=0.33\textwidth]{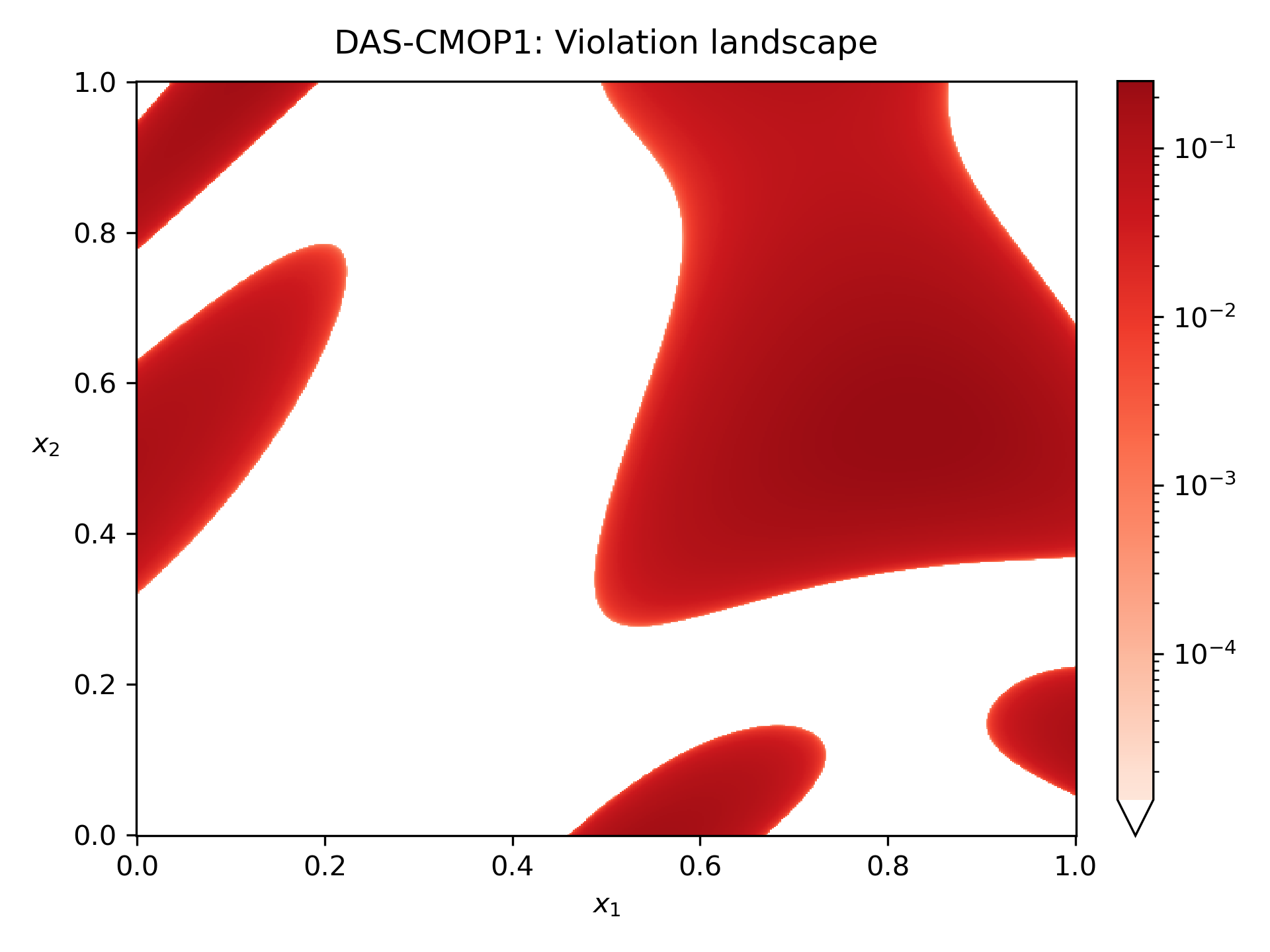}%
    \label{fig:dascmop1_vl}}
    \hfil
    \subfloat[DAS-CMOP1:\\ Problem landscape]{\includegraphics[clip, trim={0, 10pt, 0, 25pt}, width=0.33\textwidth]{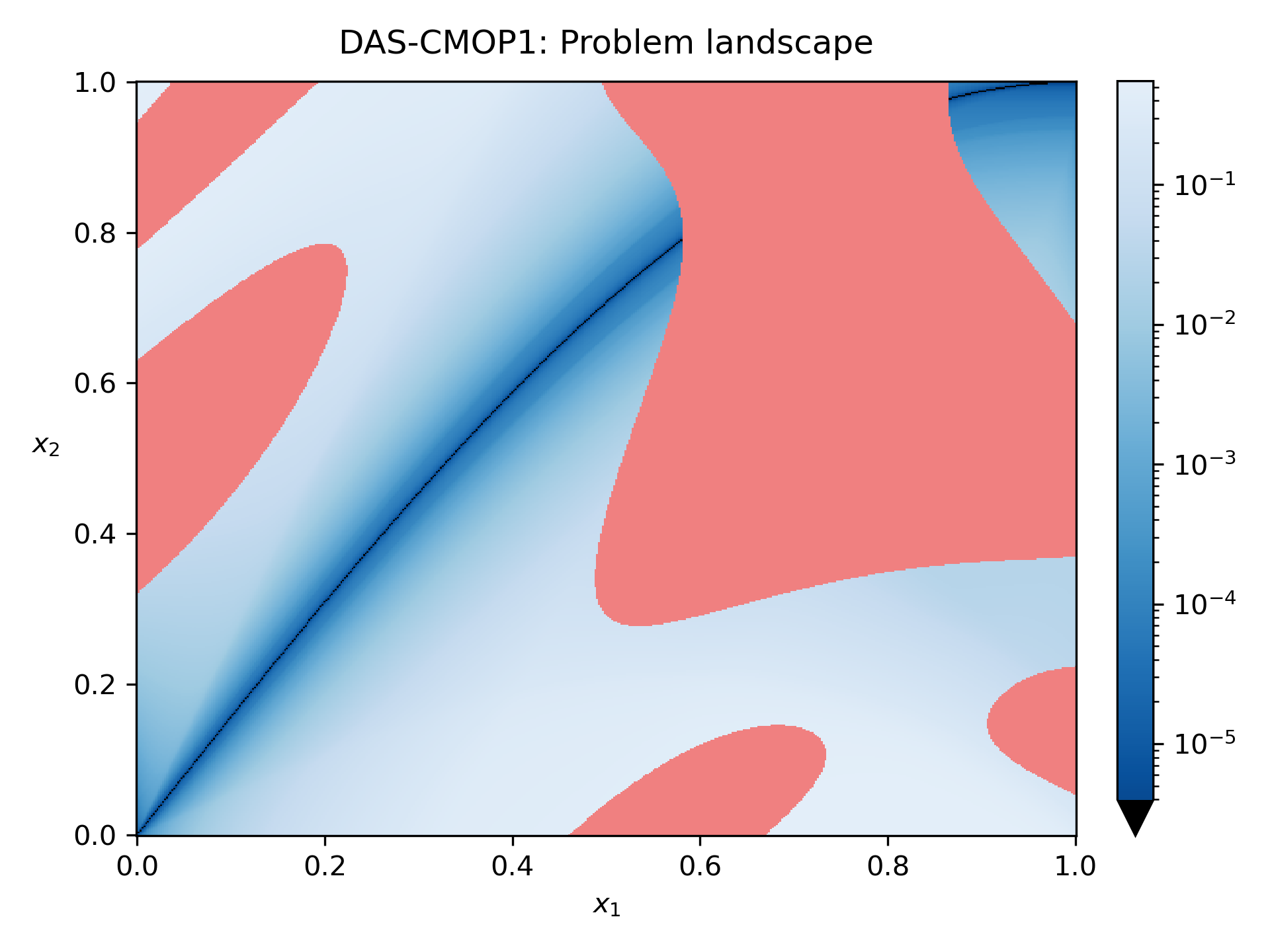}%
    \label{fig:dascmop1_pl}}
    
    \subfloat[MW7:\\ Dominance ratio]{\includegraphics[clip, trim={0, 10pt, 0, 25pt}, width=0.33\textwidth]{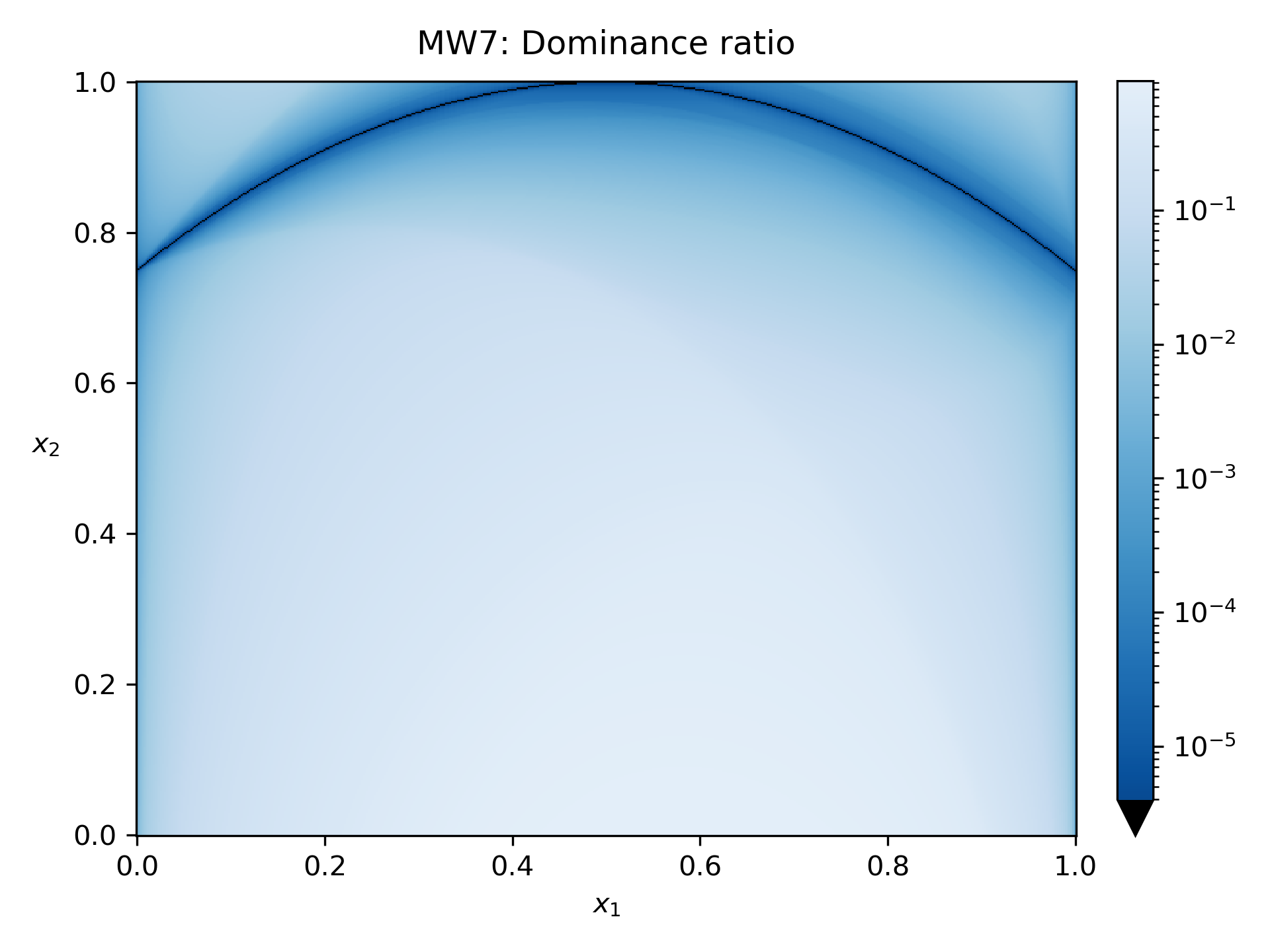}%
    \label{fig:mw7_fl}}
    \hfil
    \subfloat[MW7:\\ Violation landscape]{\includegraphics[clip, trim={0, 10pt, 0, 25pt}, width=0.33\textwidth]{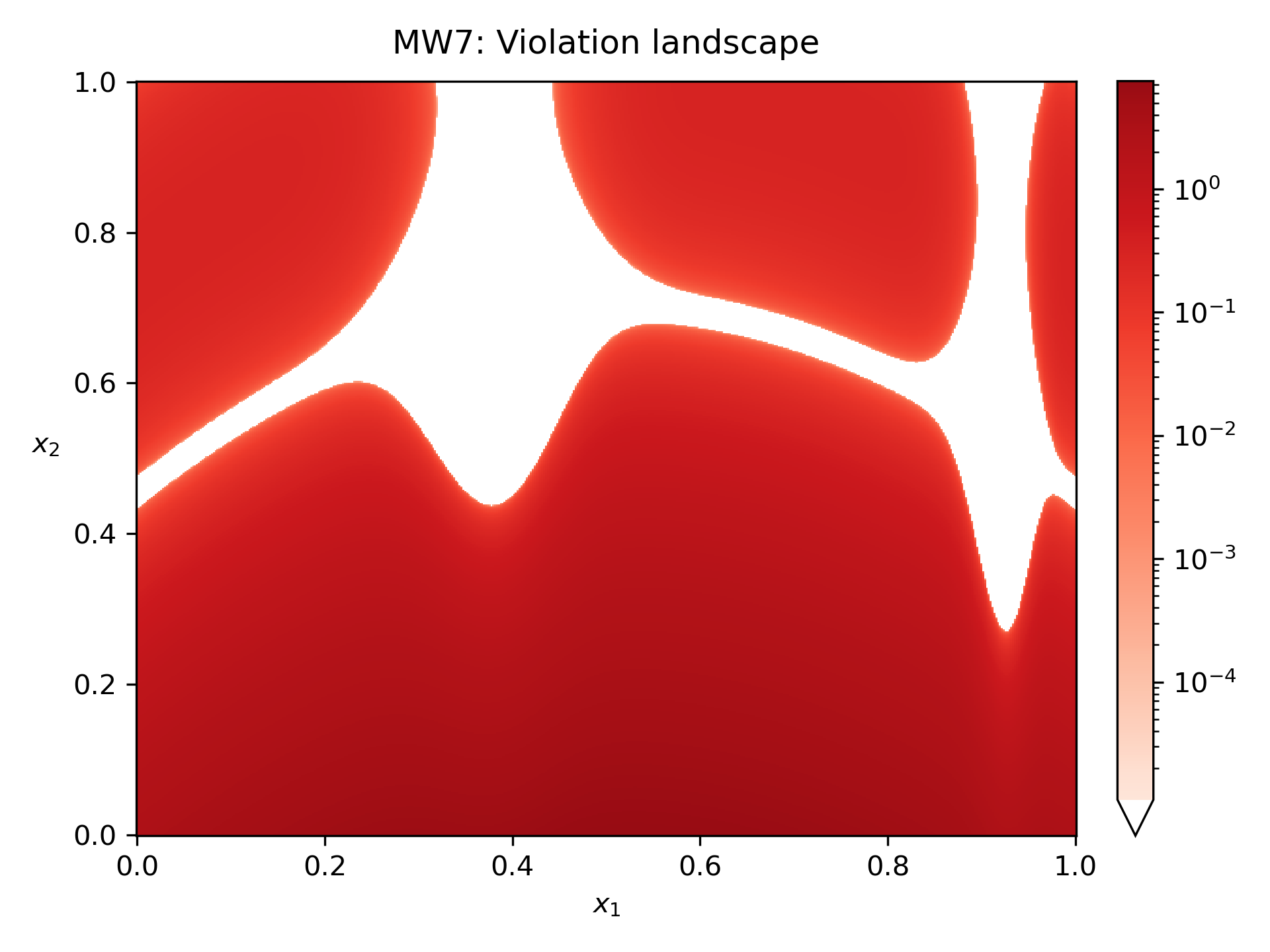}%
    \label{fig:mw7_vl}}
    \hfil
    \subfloat[MW7:\\ Problem landscape]{\includegraphics[clip, trim={0, 10pt, 0, 25pt}, width=0.33\textwidth]{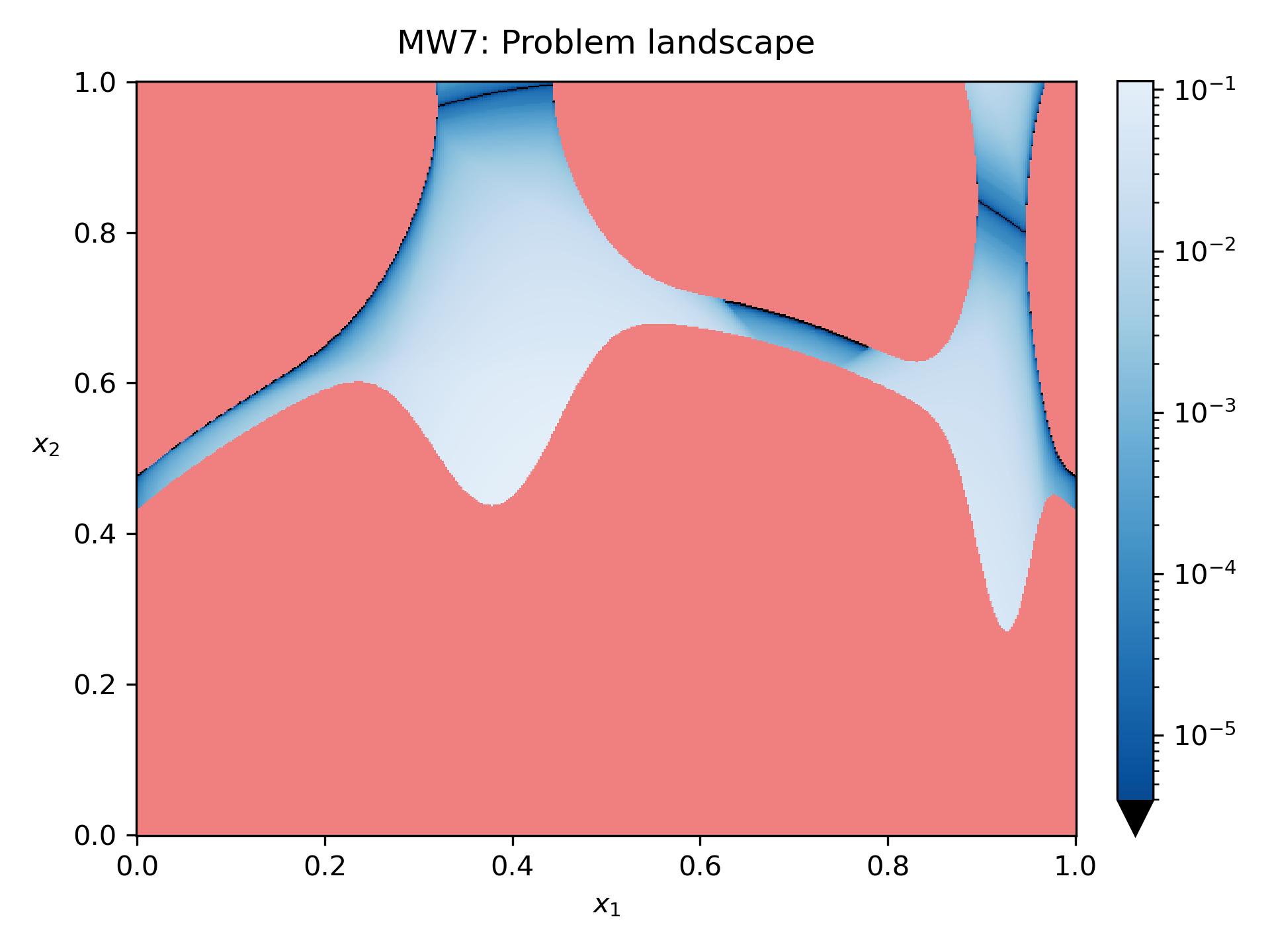}%
    \label{fig:mw7_pl}}
    
    \caption{Plots of the dominance ratio (the first column), the violation landscape (the second column), and the problem landscape (the third column) of three benchmark problems. The approximation of the Pareto-optimal set is depicted in black in dominance ratio and problem landscape plots.}
    \label{fig:landscapes}
\end{figure*}

Figure~\ref{fig:landscapes} illustrates problem landscapes and violation landscapes for problems C2-DTLZ2~\cite{Jain2014}, MW6~\cite{Ma2019}, DAS-CMOP1~\cite{Fan2019b} and MW7~\cite{Ma2019} with two variables and two objectives. All figures were obtained by evaluating a grid of $501 \times 501$ solutions from the search space. The first column shows the dominance ratio for each grid solution expressed as a proportion of grid solutions that dominate it~\cite{Fonseca1995}. Those solutions that are not dominated by any grid solution represent approximations of the Pareto-optimal solutions and are shown in black in the plots. The second column depicts the violation landscape in terms of constraint violation values with feasible components presented in white. The third column shows the problem landscape by visualizing the dominance ratio of the feasible regions in blue hues and the infeasible regions in pink. Again, black denotes approximations of Pareto-optimal solutions.

The Pareto-optimal set of C2-DTLZ2 without considering constraints is the horizontal line $L : x_2 = 0.5$ (Figure~\ref{fig:c2dtlz2_fl}). Similarly holds for MW6 (Figure~\ref{fig:mw6_fl}), while the Pareto-optimal sets of DAS-CMOP1 and MW7 without constraints are curves in the search space (Figures~\ref{fig:dascmop1_fl}~and~\ref{fig:mw7_fl}).

As we can see from Figure~\ref{fig:c2dtlz2_vl}, C2-DTLZ2 has three feasible components that are also the only three local minimum-violation components. The corresponding basins are separated with two vertical lines $L_1 : x_1 = 0.25$ and $L_2 : x_1 = 0.75$. All the feasible components and basins intersect with the Pareto-optimal set (Figure~\ref{fig:c2dtlz2_pl}). In other words, each basin attracts the search towards nondominated feasible solutions, making this problem easy for constraint handling.

On the other hand, MW6 consists of 35 feasible components and 72 basins (Figure~\ref{fig:mw6_vl}). Between two adjacent horizontal components, there is an additional basin that contains no feasible solutions. It is also interesting that not all the feasible components contain Pareto-optimal solutions (Figure~\ref{fig:mw6_pl}). This makes the MW6 problem harder to handle, especially for separation-based constraint handling techniques that separately handle feasible and infeasible solutions and strictly favor feasible solutions over the infeasible ones. The Pareto-optimal set is a subset of the Pareto-optimal set corresponding to MW6 without considering constraints.

The DAS-CMOP1 problem has three feasible components: a small one in the upper left corner, a medium-sized one in the upper right corner, and a large one in the middle. In contrast to C2-DTLZ2 and MW6, the feasible components have irregular shapes. The feasible components are also the only local minimum-violation components. The Pareto-optimal set is a subset of the Pareto-optimal set corresponding to DAS-CMOP1 without considering constraints.

In contrast, the Pareto-optimal set of the MW7 problem consists of two disconnected parts of the Pareto-optimal set belonging to the MW7 problem without constraints and parts of the boundary between the feasible and infeasible regions (Figure~\ref{fig:mw7_pl}). This problem has one feasible component and also only one basin of attraction (Figure~\ref{fig:mw7_vl}).

To identify and count the basins of attraction, we use the truncated Newton method~\cite{Nash1984} as a local search procedure $\mu$. This method approximates the Newton's direction by approximately solving the Newton equation using an iterative technique. In this study, the conjugate gradient technique is used as the iterative solver. Note that any other deterministic local search could be used instead and possibly result in a different number of basins. Nevertheless, the truncated Newton method proved to be efficient and reliable for identifying basins of attraction.

\section{Proposed ELA features}
\label{sec:methodology}

The ELA features presented in this study can be, based on the methodology used to derive them, categorized into four groups: space-filling design, information content, random walk and adaptive walk features. They are all summarized in Table~\ref{tab:features}.

\begin{table}
    \centering
    \caption{The proposed ELA features to characterize CMOPs categorized into four groups: space-filling design, information content, random walk, and adaptive walk. ``New'' indicates that the corresponding feature is proposed in this paper.}
    \label{tab:features}
    \begin{tabular}{lll}
        \hline
        \multicolumn{3}{l}{Space-filling design features} \\
        \hline
        $N_{\com{}}$ & Number of feasible components & New \\
        $\com{min}$ & Smallest feasible component & New \\
        $\com{med}$ & Median feasible component & New \\
        $\com{max}$ & Largest feasible component & New \\
        $\mathcal{O}(\com{max})$ & Proportion of Pareto-optimal solutions in $\com{max}$ & New \\
        $\com{opt}$ & Size of the ``optimal'' feasible component & New \\
        $\rho_{\mathrm{F}}$ & Feasibility ratio & \cite{Picard2021} \\
        $\rho_{\mathrm{min}}$ & Minimum correlation & \cite{Malan2015}$^a$ \\
        $\rho_{\mathrm{max}}$ & Maximum correlation & \cite{Malan2015}$^a$ \\
        $\rhops$ & Proportion of boundary Pareto-optimal solutions & New\\
        \hline
        \multicolumn{3}{l}{Information content features}\\
        \hline
        $\hmax$ & Maximum information content & \cite{Munoz2015b}$^b$ \\
        $\varepsilon_s$ & Settling sensitivity & \cite{Munoz2015b}$^b$ \\
        $M_0$ & Initial partial information & \cite{Munoz2015b}$^b$ \\
        \hline
        \multicolumn{3}{l}{Random walk features}\\
        \hline
        $\rhof{min}$ & Minimum ratio of feasible boundary crossings & \cite{Picard2021} \\
        $\rhof{med}$ & Median ratio of feasible boundary crossings & \cite{Picard2021} \\
        $\rhof{max}$ & Maximum ratio of feasible boundary crossings & \cite{Picard2021} \\
        \hline
        \multicolumn{3}{l}{Adaptive walk features}\\
        \hline
        $N_{\basin{}}$ & Number of basins & \cite{Kerschke2019}$^b$ \\
        $\basin{min}$ & Smallest basin & New \\
        $\basin{med}$ & Median basin & New \\
        $\basin{max}$ & Largest basin & New \\
        $\basinf{min}$ & Smallest feasible basin & New \\
        $\basinf{med}$ & Median feasible basin & New \\
        $\basinf{max}$ & Largest feasible basin & New \\
        $\cup \mathcal{B}_\mathrm{F}$ & Proportion of feasible basins & New \\
        $v(\basin{})_{\mathrm{med}}$ & Median constraint violation over all basins & New \\
        $v(\basin{})_{\mathrm{max}}$ & Maximum constraint violation of all basins & New \\
        $v(\basin{max})$ & Constraint violation of $\basin{max}$ & New \\
        $\mathcal{O}(\basin{max})$ & Proportion of Pareto-optimal solutions in $\basin{max}$ & New \\
        $\basin{opt}$ & Size of the ``optimal'' basin & New \\
        \hline
        \multicolumn{3}{l}{$^a$For the first time applied in multiobjective optimization in our work.} \\
        \multicolumn{3}{l}{$^b$For the first time applied on the violation landscape in our work.} \\
    \end{tabular}
\end{table}

\subsection{Space-filling design features} 
\label{sec:random_sampling_features}

The space-filling design approach employed in this paper can be described with the following three steps:
\begin{enumerate}
    \item Generate an initial sample $X_\mathrm{S} \subseteq S$ of solutions following a selected space-filling design and derive feasible solutions.
    \item Cluster feasible solutions to obtain approximations for feasible components. 
    \item Find nondominated solutions among the obtained feasible solutions.
\end{enumerate}

Following this approach, various features can be derived to characterize feasible components. The first feature of this group is the number of feasible components, $N_{\com{}}$, and it is estimated as the total number of the obtained clusters. Next, the proportion of solutions from $X_\mathrm{S}$ that form a particular cluster is used to measure the size of the corresponding feasible component. Three features were derived with this respect: the smallest feasible component, $\com{min}$, the median feasible component, $\com{med}$, and the largest feasible component, $\com{max}$. 

The following two features considered within this group express the ``optimality'' of feasible components. The first feature reflects the ``optimality'' of the largest feasible component, $\mathcal{O}(\com{max})$, and it is expressed as the proportion of nondominated feasible solutions in the largest cluster. The second feature is the size of the ``optimal'' feasible component, $\com{opt}$. It is approximated as the proportion of all solutions from the initial sample that form a cluster with the maximum number of nondominated feasible solutions.

The proportion of feasible solutions in the initial sample is used to estimate the feasibility ratio, $\rho_{\mathrm{F}}$, of the corresponding CMOP. 

In addition, two features express the correlations between the objectives and constraints. They are the minimum and maximum correlations among objectives and the overall constraint violation and are denoted by $\rho_{\mathrm{min}}$ and $\rho_{\mathrm{max}}$, respectively.

Finally, the proportion of Pareto-optimal solutions on the feasible region boundary is approximated as the proportion of nondominated solutions on the boundary of one of the obtained clusters and is denoted by $\rhops$.

\subsection{Information content features} 
\label{sec:information_content_features}

The information content approach was originally used to analyze a sequence of fitness values obtained by sorting a sample of solutions and quantifies the ``information content'' of a fitness landscape (smoothness, ruggedness, and neutrality). We extend this approach to the violation landscape. Instead of analyzing fitness sequences, we analyze sequences of overall constraint violation values. 

The approach follows the description provided in~\cite{Munoz2015b} and it is here presented only briefly. The interested reader can find a detailed explanation in the original paper. First, a sample of solutions $X_\mathrm{I} = \{x^{(1)}, \dots, x^{(n)}\}$ is generated following a space-filling design. The generated solutions from $X_\mathrm{I}$ are sorted in a sequence as follows: The first solution is selected at random, while each subsequent solution is selected as the nearest solution to the current one. Then, a new sequence, $\phi(\lambda) = \{\phi^{(1)}(\lambda), \dots, \phi^{(n-1)}(\lambda)\}$, is generated from the original one following the rule
\begin{equation}
    \phi^{(i)}(\lambda) = 
    \begin{cases}
       \searrow & \quad \text{if } \frac{\Delta v^{(i)}}{\norm{\Delta x^{(i)}}} < - \lambda \\
       \rightarrow & \quad \text{if } \frac{\abs{\Delta v^{(i)}}}{\norm{\Delta x^{(i)}}} \leq \;\;\, \lambda \\
       \nearrow & \quad \text{if } \frac{\Delta v^{(i)}}{\norm{\Delta x^{(i)}}} > \;\;\, \lambda \\
     \end{cases}
\end{equation}
where $\lambda > 0$, $\Delta x^{(i)} = x^{(i + 1)} - x^{(i)}$, $\Delta v^{(i)} = v^{(i + 1)} - v^{(i)}$ and $v^{(i)} = v(x^{(i)})$. In more detail, the symbol $\searrow$ indicates a decrease of the overall constraint violation, $\rightarrow$ a neutral area within the threshold $\lambda$, and $\nearrow$ an increase of the overall constraint violation. In particular, two consecutive symbols compose a block, $ab$, where $a, b \in \{ \searrow, \rightarrow, \nearrow \}$, which represents an object in the violation landscape, i.e., a slope, peak, etc. For example, $\nearrow\, \nearrow$ indicates a constant increase of the overall constraint violation, while $\searrow\, \nearrow$ indicates a bottom. In particular, two identical consecutive symbols represent a smooth area in the violation landscape, while two different consecutive symbols indicate a rugged area.

Finally, the information content of $X_\mathrm{I}$ is defined as the Shannon entropy of the distribution for the blocks $ab$: 
\begin{equation}
H(\lambda) = - \sum_{a \neq b} p_{ab} \log_6 p_{ab}
\end{equation}
where $p_{ab}$ is the probability of finding one of possible blocks $ab$, where $a \neq b$, within the sequence $\phi(\lambda)$. The logarithm base is six as this is the number of different blocks where $a \neq b$. The main intuition behind information content is that a large value of $H(\lambda)$ is an indication of perpetual changes in the violation landscape and therefore its ruggedness. Another important measure is the partial information content defined as $M(\lambda) = \abs{\phi'(\lambda)} / (n - 1)$, where $\phi'(\lambda)$ is a sequence derived from $\phi(\lambda)$ by removing all $\rightarrow$ and repeated symbols.

From $H(\lambda)$ and $M(\lambda)$, various features can be derived representing smoothness of the violation landscape. In this work, we study three features often used in the literature~\cite{Munoz2015a}: the maximum information content $H_{\mathrm{max}} = \max_{\lambda}\{H(\lambda)\} $, the settling sensitivity $\varepsilon_S = \log_{10} (\min_{\lambda}\{\lambda \mid H(\lambda) < 0.05\})$, and the initial partial information $M_0 = M(0)$. Features $H_{\mathrm{max}}$ and $M_0$ express the violation landscape smoothness (or ruggedness). Smooth landscapes have small values of $H_{\mathrm{max}}$ and $M_0$, and vice versa. The feature $\varepsilon_S$, on the other hand, represents the maximum change of the overall constraint violation within the original sequence.

\subsection{Random walk features} 
\label{sec:random_walk_features}

A random walk through the problem search space is a well-known and often used ELA technique in single-objective optimization. A walk is an ordered sequence of solutions $X_\mathrm{R} = \{x^{(1)}, \dots, x^{(n)}\}$ such that $x_i \in \mathcal{N}(x^{(i-1)},\delta)$ for all $i \in \{2, \dots, n\}$. During a random walk, there is no particular criterion to select the neighboring solution at each step, i.e., a random neighbor is selected~\cite{Liefooghe2019}.

For constrained single-objective optimization, a ratio of feasible boundary crossings, $\rho_{\partial F}$, was introduced in~\cite{Malan2015}. This feature quantifies the number of boundary crossings from feasible to infeasible space and vice versa, as encountered by a random walk through the search space. Given a sequence $X_\mathrm{R}$ generated from a random walk, a binary sequence is derived $B=\{b^{(1)}, \dots, b^{(n)}\}$ such that $b_i=0$ iff $x_i$ is a feasible solution. The feature $\rho_{\partial F}$ is defined as the proportion of steps in the random walk that cross the boundary of the feasible region:
\begin{equation}
    \rho_{\partial F} = \frac{1}{n - 1} \sum_{i=2}^{n} \abs{b^{(i)} - b^{(i-1)}}.
\end{equation}

Since the ratio of feasible boundary crossings does not consider the objective values, it is a feature characterizing violation landscapes. Therefore, it can be used in multiobjective optimization without any modifications~\cite{Picard2021}.

To provide a more robust measure, \cite{Malan2015} defines $\rho_{\partial F}$ as the average value over $N$ independent random walks. However, this aggregate value provides no local information at all. To avoid this shortcoming, we also consider the maximum and minimum values of $\rho_{\partial F}$ observed in $N$ random walks. The resulting features are: the minimum ratio of feasible boundary crossings $(\rho_{\partial F})_{\mathrm{min}}$, the median ratio of feasible boundary crossings $(\rho_{\partial F})_{\mathrm{med}}$, and the maximum ratio of feasible boundary crossings $(\rho_{\partial F})_{\mathrm{max}}.$ A value of $(\rho_{\partial F})_{\mathrm{min}}$ close to zero indicates that there exists at least one larger feasible (or infeasible) region. Similarly, a large value of $(\rho_{\partial F})_{\mathrm{max}}$ indicates that there is an area in the violation landscape with a heavily dissected feasible component.

\subsection{Adaptive walk features} 
\label{sec:adaptiwe_walk_features}

In contrast to random walks, an improving neighbor is selected at each step of an adaptive walk, as a local search would do. In unconstrained single-objective optimization, adaptive walks are often used to derive various features to characterize basins of attraction of the fitness landscape. In this paper, we extend these approaches to the violation landscape to quantitatively describe its basins of attraction. 

The approach considered in this study is similar to the technique described in~\cite{Kerschke2019}. However, various novel features specialized for CMOPs are introduced here for the first time. The approach consists of four steps:
\begin{enumerate}
     \item Generate an initial sample $X_\mathrm{A} \subseteq S$ of solutions following a selected space-filling design.
     \item Perform a local search for each solution $x \in X_\mathrm{A}$ to obtain local minimum-violation solutions. 
     \item Cluster local minimum-violation solutions to obtain approximations for local minimum-violation components and basins of attraction.
     \item Find nondominated feasible solutions among the obtained local minimum-violation solutions. 
\end{enumerate}

Using the above approach, we can derive various features to characterize basins of attraction of the violation landscape. The resulting total number of the found clusters provides an approximation for the number of basins, $N_{\basin{}}$, the first feature of this group. The proportion of all solutions from the initial sample that converge to a specific cluster is used to measure the size of the corresponding basin. The resulting features are: the smallest basin, $\basin{min}$, the median basin, $\basin{med}$, and the largest basin, $\basin{max}$. Similarly, we derive three equivalent features for feasible basins (basins containing feasible solutions): the smallest feasible basin, $\basinf{min}$, the median feasible basin, $\basinf{med}$, and the largest feasible basin, $\basinf{max}$. In addition, the proportion of all solutions that converge to one of the feasible basins reflects the size of the union of all feasible basins, $\cup \mathcal{B}_\mathrm{F}$.

Next, the minimum constraint violation value of all local minimum-violation solutions in a cluster approximates the constraint violation of the corresponding basin defined as $v(B) = \inf \{v(x) \mid x \in \basin{}\}$. Three features are considered: the median constraint violation over all basins, $v(\basin{})_{\mathrm{med}}$, the maximum constraint violation of all basins, $v(\basin{})_{\mathrm{max}}$, and the constraint violation of the largest basin, $v(\basin{max})$.

Finally, the last two features considered in this group indicate the ``optimality'' of basins of attraction. The first feature is the ``optimality'' of the largest basin, $\mathcal{O}(\basin{max})$, expressed as the number of nondominated feasible solutions in the largest cluster. The second feature is the size of the ``optimal'' basin, $\basin{opt}$, expressed as the proportion of solutions that converge to the cluster with the maximum number of nondominated feasible solutions.

\section{Experimental analysis} 
\label{sec:experiments}

In this section, we first introduce the test suites chosen for the experiments. Next, we discuss the experimental setup, including the specific techniques and parameter settings used to derive the ELA features. The derived features are then used to evaluate the existing test suites. Additionally, we present the sensitivity analysis of the parameters used for feature extraction, and finally, we comment on the scalability of the applied space-filling design techniques.

\subsection{Test suites} 
\label{sec:test_suites}

The most notable artificial test suites of CMOPs used to asses the performance of constrained multiobjective optimization algorithms are CTP~\cite{Deb2001}, CF~\cite{Zhang2008}, C-DTLZ~\cite{Jain2014}, NCTP~\cite{Li2016}, DC-DTLZ~\cite{Li2019}, LIR-CMOP~\cite{Fan2019a}, DAS-CMOP~\cite{Fan2019b}, and MW~\cite{Ma2019}. 

All the mentioned test suites consist of artificially designed CMOPs and are not derived from real-world applications. To overcome this weakness, a novel suite named RCM was proposed in~\cite{Kumar2021}. The RCM suite collects 50 real-world optimization problems based on physical models, including problems from mechanical design, chemical engineering, power electronics, etc. The problem instances come in various dimensions and numbers of objectives. After excluding all the problems containing discrete variables and those with more than five variables, 11 RCM problems remained for our experimental analysis and are presented in Table~\ref{tab:rcm}. As we can see, ten of the selected RCM problems have two objectives while the reaming one has five objectives. The number of constraints varies from one to eight, and all the constraints are inequalities.

\begin{table}
    \centering
    \caption{Characteristics of the selected RCM problems: dimension of the search space $D$, number of objectives $M$, and number of constraints $I$ (all constraints are inequalities).}
    \label{tab:rcm}
    \begin{tabular}{lllll}
        \hline
        Test problem & Name & $D$ & $M$ & $I$ \\
        \hline
        RCM2 & Vibrating platform design & 5 & 2 & 5 \\
        RCM3 & Two bar truss design & 3 & 2 & 3 \\
        RCM4 & Welded beam design & 4 & 2 & 4 \\
        RCM5 & Disc brake design & 4 & 2 & 4 \\
        RCM10 & Two bar plane truss design & 2 & 2 & 2 \\
        RCM11 & Water resources management & 3 & 5 & 7 \\
        RCM12 & Simply supported beam design & 4 & 2 & 1 \\
        RCM14 & Multiple disk clutch brake design & 5 & 2 & 8 \\
        RCM16 & Cantilever beam design & 2 & 2 & 2 \\
        RCM18 & Front rail design & 3 & 2 & 3 \\
        RCM20 & Hydro-static thrust bearing design & 4 & 2 & 7 \\
        \hline
    \end{tabular}
\end{table}

Note that an additional real-world multiobjective optimization problem suite including eight CMOPs was proposed in~\cite{Tanabe20}. Nevertheless, all relevant CMOPs from this suite are also included in the RCM suite, and it was therefore omitted from this study.

The basic characteristics of the test suites used in this study are summarized in Table~\ref{tab:suites}.

\begin{table}
    \centering
    \caption{Characteristics of test suites: number of problems, dimension of the search space $D$, number of objectives $M$, and number of constraints $I$ (all constraints are inequalities). The characteristics of the selected RCM problems (see Table~\ref{tab:rcm}) are shown in parentheses.}
    \label{tab:suites}
    \begin{tabular}{lllll}
        \hline
        Test suite & \#problems & $D$ & $M$ & $I$ \\
        \hline
        CTP~\cite{Deb2001} & \phantom{0}8 & * & 2 & 2, 3 \\
        CF~\cite{Zhang2008} & 10 & * & 2, 3 & 1, 2 \\
        C-DTLZ~\cite{Jain2014} & \phantom{0}6 & * & * & 1, * \\
        NCTP~\cite{Li2016} & 18 & * & 2 & 1, 2 \\
        DC-DTLZ~\cite{Li2019} & \phantom{0}6 & * & * & 1, * \\
        DAS-CMOP~\cite{Fan2019b} & \phantom{0}9 & * & 2, 3 & 7, 11 \\
        LIR-CMOP~\cite{Fan2019a} & 14 & * & 2, 3 & 2, 3 \\ 
        MW~\cite{Ma2019} & 14 & * & 2, * & 1--4 \\
        RCM~\cite{Kumar2021} & 50 (11) & 2--34 (2--5) & 2, 5 & 1--29 (1--8) \\
        \hline
        \multicolumn{5}{l}{*Scalable parameter.}\\
    \end{tabular}
\end{table}

\subsection{Experimental setup} 
\label{sec:experimental_setup}

The proposed features are demonstrated on the test suites listed in Section~\ref{sec:test_suites}. In particular, bi-objective C-DTLZ and DC-DTLZ problems were considered with the default number of constraints. Additionally, a difficulty triplet of (0.5, 0.5, 0.5) was used for the DAS-CMOP suite as this is by far the most frequently used difficulty triplet in the literature.

Three dimensions of the search space (2, 3, 5) were considered to measure the scalability of various techniques used to derive the ELA features. For each dimension, initial sample sizes were decided based on some initial experiments and were selected as the minimum number of solutions needed for the feature values to converge. The details are presented in Table~\ref{tab:setting}. Note that because the real-world problems from the RCM test suite are not scalable, the parameter values used for artificial problems with five variables were used also for the four-dimensional RCM problems.

To produce the initial samples used by all the proposed techniques to derive ELA features, we used the space-filling design based on Latin hypercube sampling.

We used the density-based spatial clustering of applications with noise (DBSCAN)~\cite{Ester1996} as the clustering algorithm for the identification of feasible components and basins, as well as to discover nondominated solutions on the boundary of the feasible region. The following DBSCAN configuration was used in the experiments: distance metric was set to the Euclidean distance, the number of samples in a neighborhood for a point to be considered a core point was set to five. At the same time, the maximum distance between two solutions for one to be considered as in the neighborhood of the other, $\varepsilon$, was defined based on initial experiments for each dimension separately (see Table~\ref{tab:setting}).

\begin{table}
    \centering
    \caption{Parameters used in the experimental analysis: Dimension of the search space $D$, initial sample size for space-filling design $\abs{X_\mathrm{S}}$ and adaptive walk $\abs{X_\mathrm{A}}$, and maximum distance between two solutions $\varepsilon$.}
    \label{tab:setting}
    \begin{tabular}{cccc}
    \hline
    $D$ & 2 & 3 & 4, 5 \\
    $\abs{X_\mathrm{S}}$ & 25\,000 & 100\,000 & 250\,000 \\
    $\abs{X_\mathrm{A}}$ & 10\,000 & \phantom{0}25\,000 & \phantom{0}50\,000 \\
    $\varepsilon$ & 0.02 & 0.04 & 0.12 \\
    \hline
    \end{tabular}
\end{table}

To express the relationship between the objectives and constraints ($\rho_{\mathrm{min}}$ and $\rho_{\mathrm{max}}$), the Spearman's rank correlation coefficient was used.

Following the recommendation from~\cite{Munoz2015b}, the initial sample size $X_{\mathrm{I}}$ for the information content features was set to $1\,000D$.

Simple random walks~\cite{Malan2014} were employed for calculating features representing ratios of feasible boundary crossings. The number of steps $X_\text{R}$ was set to 10\,000, each with maximum step size, $\delta$, of 1\% of the range of the domain for each test problem~\cite{Malan2015}. Thirty independent runs were conducted to obtain the values for $\rhof{min}$, $\rhof{med}$ and $\rhof{max}$. 

Finally, we used the truncated Newton method~\cite{Nash1984} as a local search procedure used in adaptive walks.

\subsection{Implementation details}

All CMOPs and techniques to derive the ELA features were implemented in the Python programming language~\cite{Rossum09}. We used the \texttt{pymoo}~\cite{Blank20} implementation for CTP, DAS-CMOP and MW, while the rest of the suites was reimplemented from scratch. Next, \texttt{pyDOE}~\cite{pydoe} was used for Latin hypercube sampling, \texttt{scikit-learn}~\cite{sklearn} for DBSCAN, \texttt{pflacco}~\cite{pflacco} for the calculation of information content features, and \texttt{SciPy}~\cite{scipy} for the truncated Newton method and to calculate Spearman’s rank correlation coefficients. The rest of the functionalities, including the simple random walk, was implemented from scratch.

\subsection{Results} 
\label{sec:results}

\begin{figure*}[!t]
    \centering
    
    \subfloat[Size of the optimal feasible component, $\com{opt}$]{\includegraphics[clip, trim={0, 10pt, 0, 30pt}, width=\textwidth]{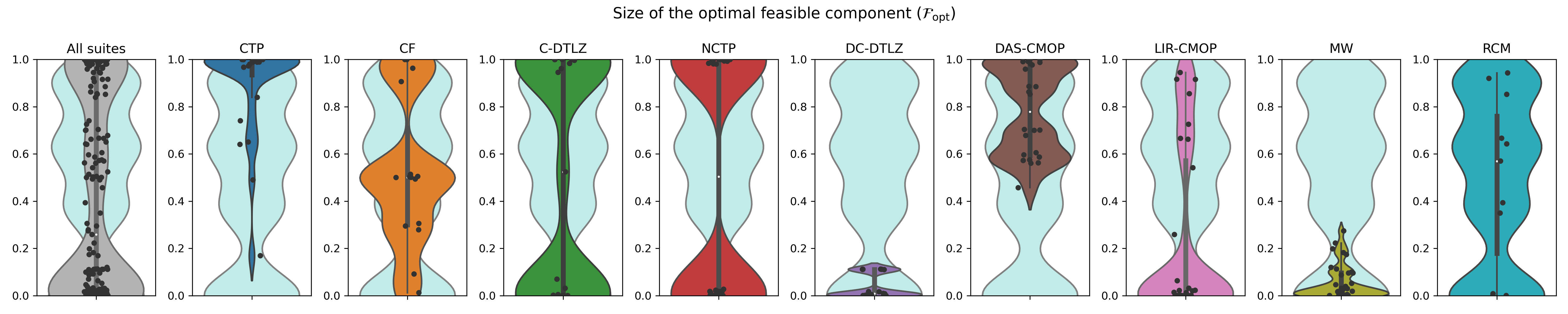}%
    \label{fig:size_opt_com}}
    
    \subfloat[Minimum correlation, $\rho_{\mathrm{min}}$]{\includegraphics[clip, trim={0, 10pt, 0, 30pt}, width=\textwidth]{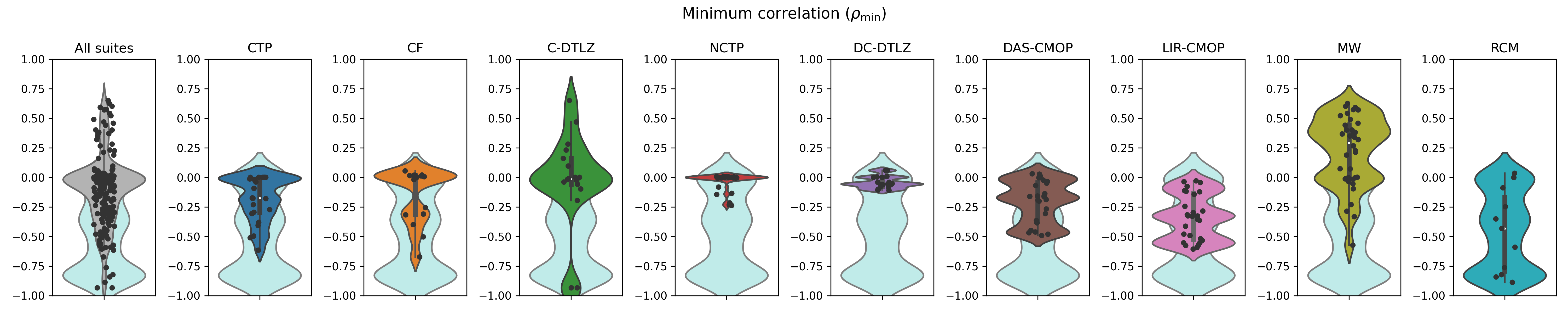}%
    \label{fig:min_corr}}
    
    \subfloat[Maximum information content, $\hmax$]{\includegraphics[clip, trim={0, 10pt, 0, 30pt}, width=\textwidth]{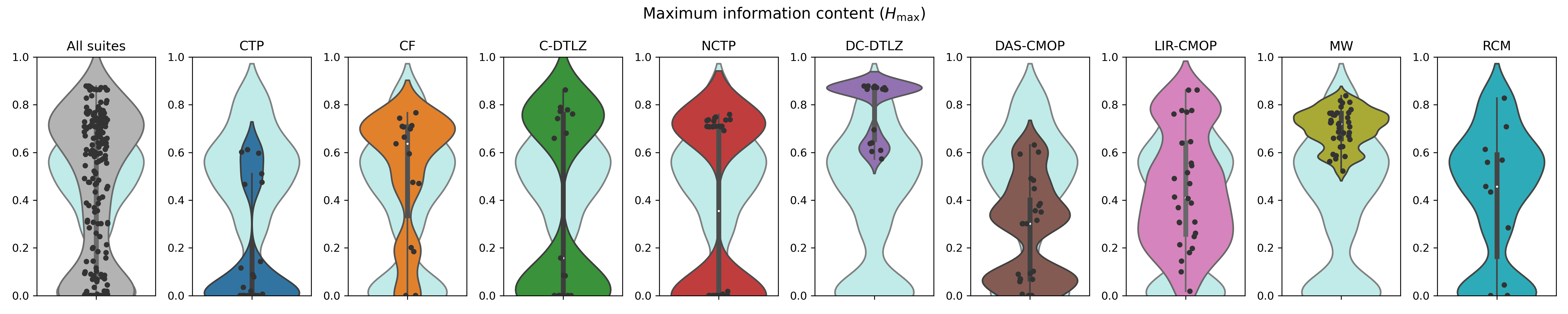}%
    \label{fig:h_max}}
    
    \subfloat[Number of basins, $N_{\basin{}}$]{\includegraphics[clip, trim={0, 10pt, 0, 30pt}, width=\textwidth]{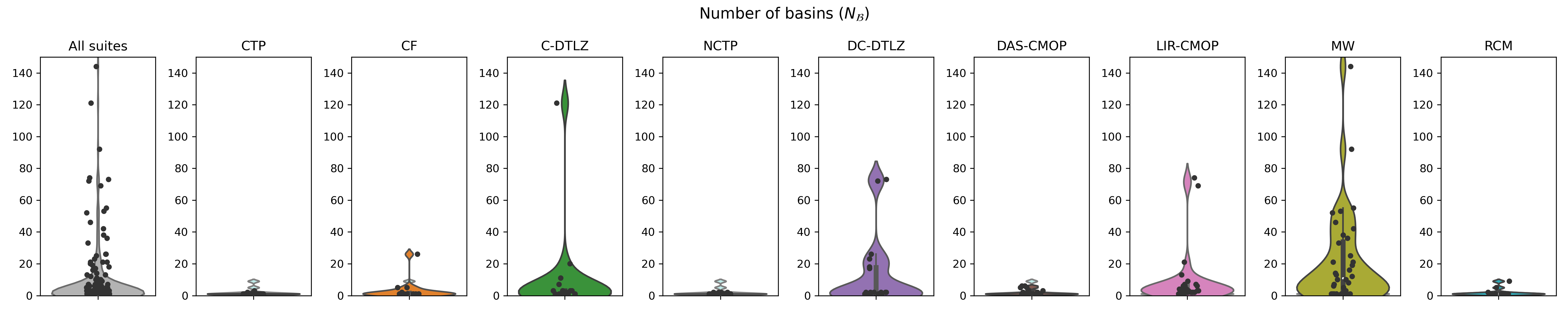}%
    \label{fig:n_basin}}
    
    \caption{Violin plots depicting selected feature distributions for the set of all the considered CMOPs (the first column) and for each suite separately (the remaining columns). The violin plot in light blue behind each suite corresponds to the RCM suite.}
    \label{fig:violin_plots_rcm}
\end{figure*}

\begin{figure*}[!t]
    \centering
    
    \subfloat[Proportion of Pareto-optimal solutions in largest feasible component, $\mathcal{O}(\com{max})$]{\includegraphics[clip, trim={0, 10pt, 0, 30pt}, width=\textwidth]{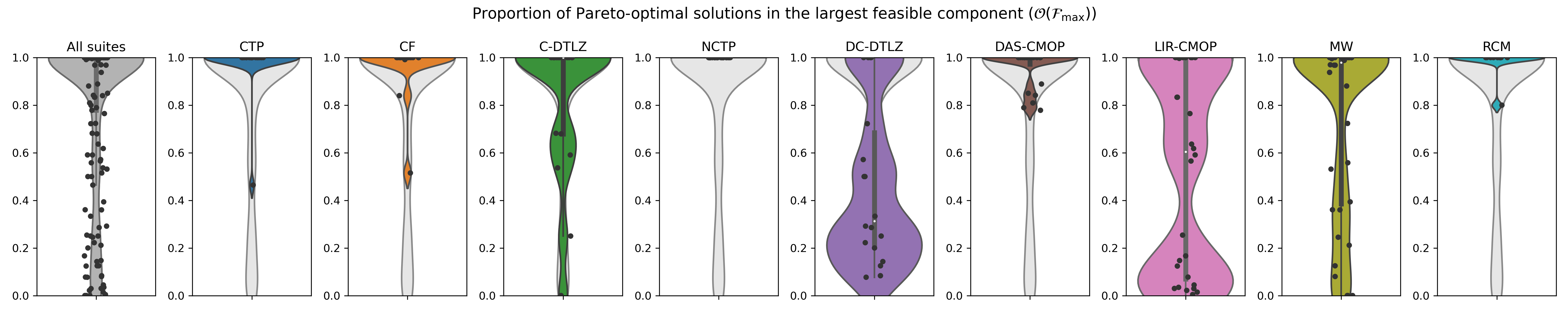}%
    \label{fig:opt_max_com}}
    
    \subfloat[Median ratio of feasible boundary crossings, $\rhof{med}$]{\includegraphics[clip, trim={0, 10pt, 0, 30pt}, width=\textwidth]{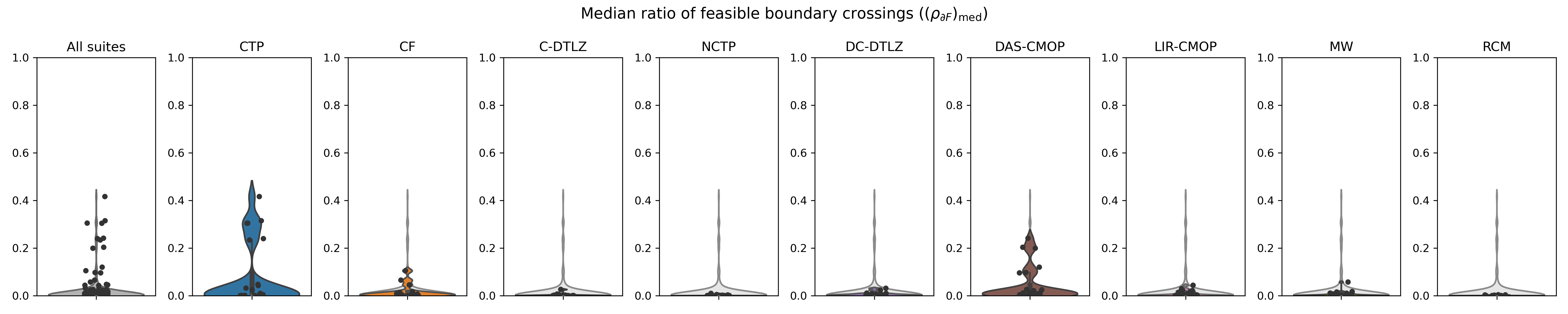}%
    \label{fig:median_rfb}}
    
    \subfloat[Proportion of Pareto-optimal solutions in largest basin, $\mathcal{O}(\basin{max})$]{\includegraphics[clip, trim={0, 10pt, 0, 30pt}, width=\textwidth]{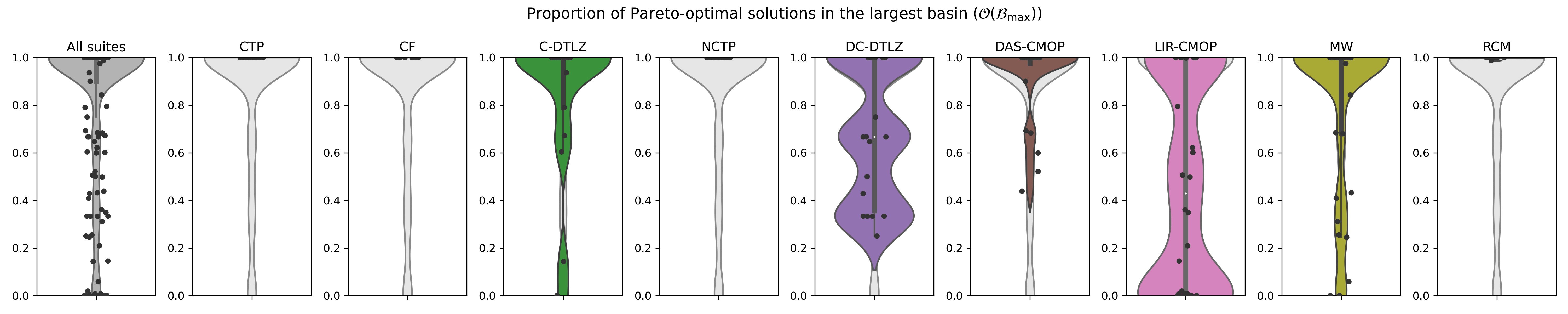}%
    \label{fig:opt_max_basin}}
    
    \caption{Violin plots depicting selected feature distributions for the set of all the considered CMOPs (the first column) and for each suite separately (the remaining columns). The violin plot in light gray behind each suite corresponds to the set of all CMOPs.}
    \label{fig:violin_plots_all}
\end{figure*}

Figures~\ref{fig:violin_plots_rcm}~and~\ref{fig:violin_plots_all} show violin plots of distributions for seven selected features $\com{opt}$, $\rho_{\mathrm{min}}$, $\hmax$, $N_{\basin{}}$, $\mathcal{O}(\com{max})$, $\rhof{med}$ and $\mathcal{O}(\basin{max})$. The first column represents the distribution for the set of all the considered CMOPs, while the rest of the columns correspond to each suite separately. Each black dot represents one problem instance. Its $y$-axis depicts the feature value, while the $x$-axis has no specific meaning and is used for better visualization only. The violin plot (colored area) approximates the probability density function for the feature distribution. For example, Figure~\ref{fig:min_corr} shows there are more problem instances in the CF suite (third column, violin plot in orange) with $\rho_{\mathrm{min}} \approx 0$ than those with $\rho_{\mathrm{min}} \approx -0.75$. In addition, there are no problem instances with $\rho_{\mathrm{min}} \approx -1$. The violin plot in light blue behind each suite corresponds to the distribution of RCM in Figure~\ref{fig:violin_plots_rcm}, and in light gray to the distribution of the set of all CMOPs in Figure~\ref{fig:violin_plots_all}. The idea is to show the coverage of RCM characteristics by the artificial test suites (Figure~\ref{fig:violin_plots_rcm}) and to expose the differences between a single test suite and the rest of the suites (Figure~\ref{fig:violin_plots_all}).

\begin{figure}[!t]
    \centering
    \subfloat[Coverage of RCM]{\includegraphics[clip, trim={0, 10pt, 0, 0}, width=0.5\textwidth]{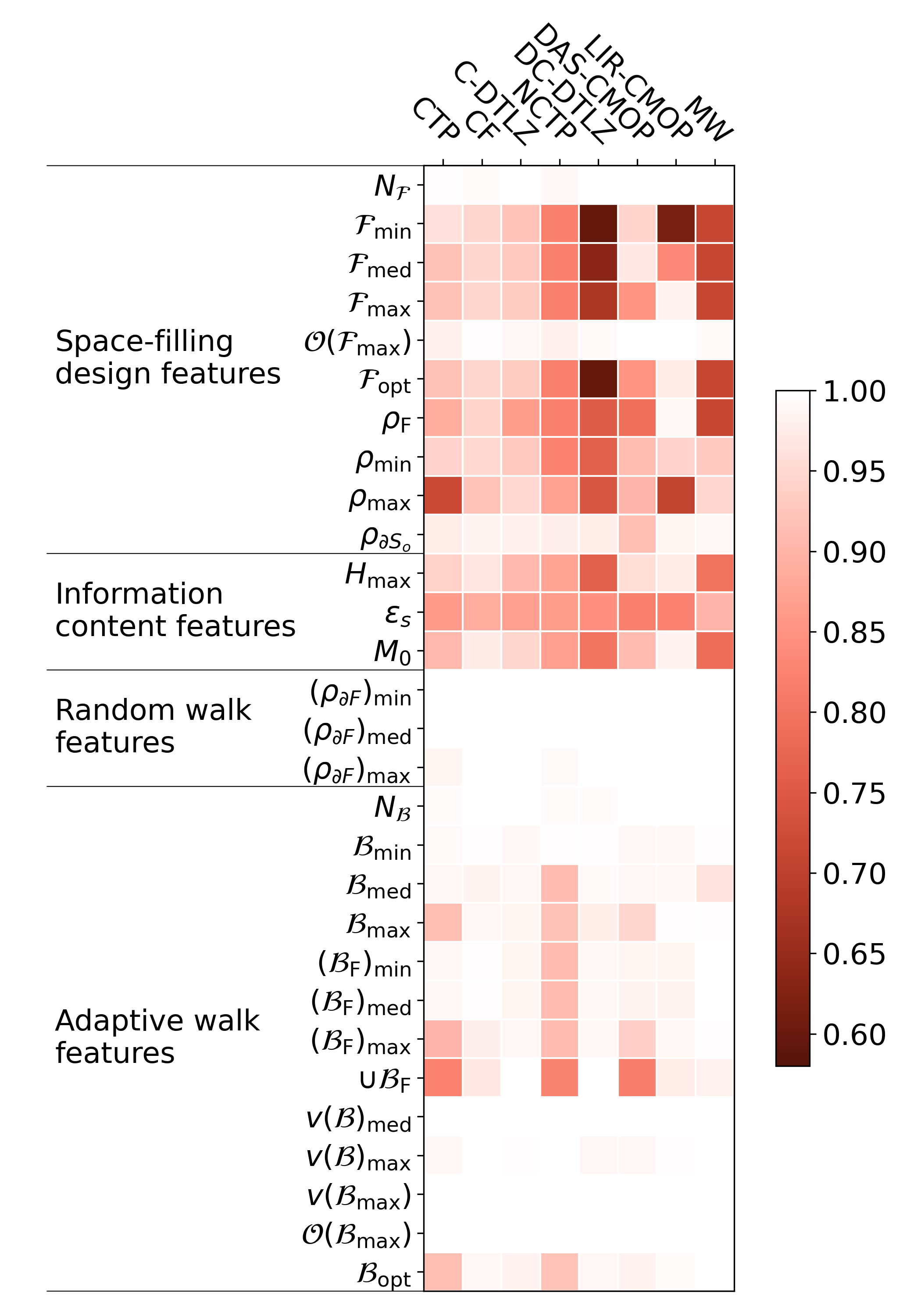}%
    \label{fig:heatmap_rcm}}
    \hfil
    \subfloat[Overall coverage]{\includegraphics[clip, trim={0, 10pt, 0, 0}, width=0.5\textwidth]{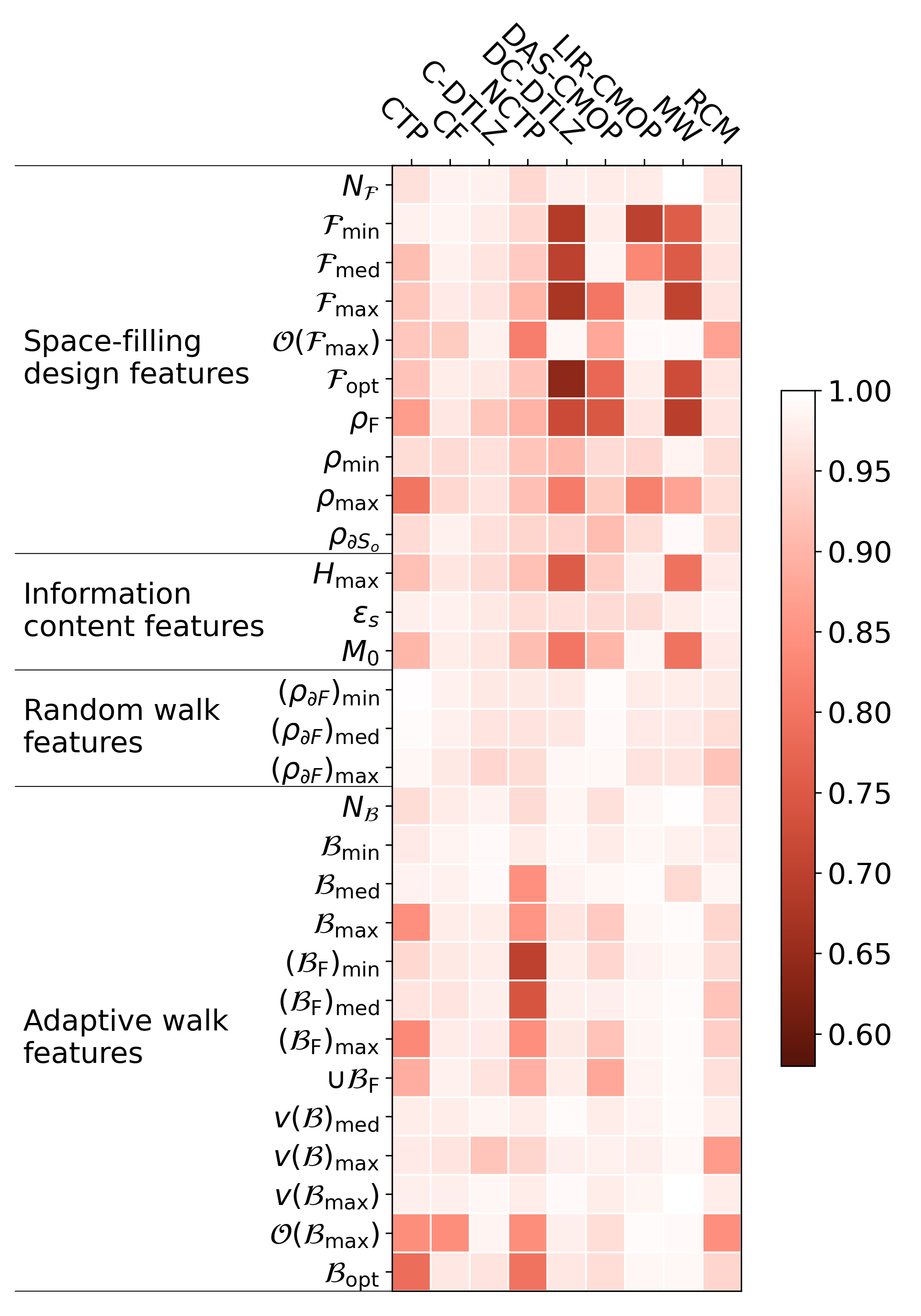}%
    \label{fig:heatmap_all}}
    \caption{Heat maps showing the coverage of characteristics present (a) in the RCM suite and (b) in the set of all the considered CMOPs.}
    \label{fig:heatmap}
\end{figure}

In addition, Figure~\ref{fig:heatmap} shows two heat maps depicting the coverage of RCM characteristics by the artificial test suites, and the coverage of characteristics present in the set of all the considered CMOPs. The coverage is expressed as one minus the mean value of all distances between the feature values from the RCM suite (or the set of all CMOPs) and the nearest feature value from a selected single suite. Note that feature values need to be normalized to $[0,1]$ before calculating the coverage values. A coverage value near one indicates excellent coverage. This means that for each feature value from RCM (or the set of all CMOPs), there exists at least one problem from the selected single suite whose feature value is close to that one from RCM (or the set of all CMOPs). In Figure~\ref{fig:heatmap}, a lighter color indicates that a feature value is well covered by a test suite, and vice versa. For example, in Figure~\ref{fig:heatmap_rcm}, $N_{\com{}}$ (first row) as represented by RCM is well covered by all the artificial suites, while $\com{max}$ (fourth row) is poorly covered, especially by DC-DTLZ and MW.

\subsection{Test suites evaluation}

According to~\cite{Bartz-Beielstein2020}, a good benchmark suite should ``include the difficulties that are typical of real world instances of the problem class under investigation''. To evaluate whether considered artificial test suites contain the characteristics observed in real-world CMOPs, we use 11 continuous and low-dimensional problems from the RCM suite and the coverage metric proposed in Section~\ref{sec:results}. It is to be noted that the low number of considered real-world problems limits the generality of the conclusions that can be drawn from these results. For example, we cannot claim that some problem characteristics that are missing from these 11 problems are in fact unrealistic. Nevertheless, we can identify which characteristics of these real-world problems are not present in the artificial problems.

The results show that several features characterizing feasible components are not adequately covered by the studied artificial test suites (Figure~\ref{fig:heatmap_rcm}, features from $N_{\com{}}$ to $\com{opt}$). Especially the recently proposed artificial test suites NCTP, DC-DTLZ, LIR-CMOP, and MW insufficiently represent problem characteristics of the studied RCM problems. For example, many of the considered RCM problems have only one feasible component, while recent artificial test problems have many feasible components. As a consequence, the RCM problems result in feasible components of any size, while, for example, the MW suite contains problems with rather small feasible components only. This shortcoming of the artificial test suites can be observed nicely for $\com{opt}$ (Figure~\ref{fig:size_opt_com}).

Similarly, the existing artificial test suites fail to satisfactorily represent the conflicts between the objectives and the overall constraint violation. As we can see in Figure~\ref{fig:min_corr}, strong negative correlations, $\rho_{\mathrm{min}} \in [-1.00, -0.75]$, observed in the studied RCM problems, are found in the C-DTLZ suite only. We can argue that a general multiobjective optimization algorithm would most likely struggle to maintain good convergence and diversity when the objectives and the overall constraint violation are strongly negatively correlated. However, our findings show that the studied artificial test suites do not cover this particular characteristic well.

The results also indicate that the feasibility ratio and information content features are poorly covered by the artificial test suites except for CFs and LIR-CMOPs (Figure~\ref{fig:heatmap_rcm}, features $\rho_{\mathrm{F}}$, $\hmax$, $\varepsilon_s$ and $M_0$). While the RCM problems take a full range of feature values, the artificial test problems have often extremely small or large feasible ratios and extremely smooth or very rugged violation landscapes (Figure~\ref{fig:h_max}).

On the other hand, the features related to basins of attraction are mostly covered well by the studied artificial test suites (Figure~\ref{fig:heatmap_rcm}, features from $N_{\basin{}}$ to $\basin{opt}$). The suites that are not diverse enough considering the basin-related features are CF, NCTP, and DAS-CMOP. Additionally, the ratios of feasible boundary crossings are well covered by the considered artificial test suites, while $\rhops$ is very well covered by all the artificial test suites except for DAS-CMOP.

Finally, the studied artificial test suites have many basin-related characteristics that cannot be observed in the selected RCM problems. While the considered RCM problems often contain only one basin where all Pareto-optimal solutions are located, many recently proposed artificial test problems have several basins, and the Pareto-optimal solutions are spread among multiple basins (Figures~\ref{fig:n_basin},~\ref{fig:opt_max_com}~and~\ref{fig:opt_max_basin}). This is in line with the recent development in constrained multiobjective optimization where the primary source of complexity is severe violation multimodality (e.g., Figure~\ref{fig:mw6_pl})~\cite{Filipic21}. However, our study suggests that other relevant aspects, such as the sizes of feasible components and basins of attraction, negatively correlated objectives and the overall constraint violation, diverse ruggedness and feasibility ratios, are frequently insufficiently addressed by the literature.

An additional analysis is devoted to measuring whether the considered test suites contain a wide variety of problems with different characteristics. This is done in the same way as the comparison with RCM, except that in this case the test suites are contrasted against the set of all CMOPs. The result are shown in Figure~\ref{fig:heatmap_all}.

Interestingly, the results are not drastically different from the previous ones, indicating that RCM contains problems with various characteristics. For example, the coverage of most of the space-filling design and information content features is very similar (Figure~\ref{fig:heatmap_all}, features from $N_{\com{}}$ to $M_0$). This is also true for seven features concerning basins of attraction $\basin{med}$, $\basin{max}$, $\basinf{min}$, $\basinf{med}$, $\basinf{max}$, $\cup \mathcal{B}_\mathrm{F}$ and $\basin{opt}$, except for the NCTP suite which has significantly worse coverage when contrasted against the set of all CMOPs.

Within the space-filling design, the only true exception is $\mathcal{O}(\com{max})$. The coverage for this feature is significantly worse when comparing the suites against the set of all CMOPs. This happens because RCM does not cover $\mathcal{O}(\com{max})$ particularly well (Figure~\ref{fig:opt_max_com}), which has already been explained before. As we can see, this aspect is even more drastically manifested for $\mathcal{O}(\basin{max})$ (Figure~\ref{fig:opt_max_basin}).

Finally, the coverage of random walk features is slightly worse when the suites are compared against the set of all CMOPs (Figure~\ref{fig:heatmap_all}, features $\rhof{min}$, $\rhof{med}$ and $\rhof{max}$). This is true because RCM contains only problems with small values of these particular features. However, in general, this set of features is well covered. Almost all the problems have only small ratios of feasible boundary crossings except for few problems from CTP, CF and DAS-CMOP (Figure~\ref{fig:median_rfb}). The same observation holds for the space-filling design feature $N_{\com{}}$, and for the following adaptive walk features $N_{\basin{}}$, $\basin{min}$, $v(\basin{})_{\mathrm{med}}$, $v(\basin{})_{\mathrm{max}}$ and $v(\basin{max})$.

In summary, we can see that all the studied test suites have some advantages and some limitations. For this reason, we recommend to include CMOPs from various suites to construct a comprehensive and sound benchmark.

\subsection{Sensitivity analysis}

The crucial step in deriving space-filling design and adaptive walk features is the accurate identification of feasible components and basins of attraction, respectively. Once the feasible components and basins are identified, all other features can be unequivocally derived. On the other hand, if the clustering step incorrectly determines the feasible components and basins of attraction, the rest of the features is also affected. The identification of both feasible components and basins of attraction strongly depends on two parameters: initial sample size ($\abs{X_\mathrm{S}}$ and $\abs{X_\mathrm{A}}$) and the maximum distance between two solutions $\varepsilon$ used by the DBSCAN algorithm.

The goal is to find the smallest sample size and the smallest $\varepsilon$ that would still yield reliable results. Small sample sizes are preferred since they require fewer function evaluations. Additionally, a small $\varepsilon$ is necessary to not erroneously merge two or more disjoint feasible components (or basins of attraction). On the other hand, a too small $\varepsilon$ can result in the separation of feasible components (or basins) into two or more components (or basins). For this reason, a careful selection of these parameters is necessary.

Space-filling design features are more sensitive to parameter setting than adaptive walk features. All infeasible solutions in the space-filling design approach are discarded and used only for enumeration. On the other hand, almost all initial solutions for adaptive walk features produce local-minimum violation solutions, thus facilitating the clustering procedure. For this reason, we investigate the sensitivity of space-filling design only. The presented analysis is very similar for adaptive walk features, except that smaller initial sample sizes are required in this case.

The goal of the presented analysis is to measure the parameter sensitivity for the identification of feasible components. More specifically, we study the extraction of the number of feasible components $N_{\com{}}$ with respect to 1) different parameter values and 2) repetition of the experiments, since a stochastic approach is used to obtain the initial sample. To address 1), we test the space-filling design approach using all combinations of the following parameter values: 
\begin{equation}
    \abs{X_\mathrm{S}} \in \{10\,000, 25\,000, 50\,000, 100\,000, 250\,000, 500\,000\}
\end{equation}
and
\begin{equation}
    \varepsilon \in \{0.01, 0.02, \dots, 0.14, 0.15\}.
\end{equation}
For 2), we repeat all the experiments 30 times, each time with a different initial sample. The analysis is performed for $D=2,3$, as the correctness of the derived components can be visually checked, while this is not possible for $D=5$.

The sensitivity analysis is performed on two CMOPs: C2-DTLZ2 and DAS-CMOP1. On the one hand, C2-DTLZ2 represents a simple problem concerning feature extraction. It has three well-shaped feasible components that are spaced apart (Figure~\ref{fig:c2dtlz2_vl}). On the other hand, the feasible components of DAS-CMOP1 have irregular shapes, and the small one in the upper left corner is located close to the large one in the middle (Figure~\ref{fig:dascmop1_vl}). This is precisely what makes the identification of components hard for DBSCAN. The 2-D DAS-CMOP1 problem with two variables has three components, while the DAS-CMOP1 with three variables has five components.

\begin{figure}[!ht]
    \centering
    \subfloat[C2-DTLZ2 ($D = 2$)]{\includegraphics[height=227pt,trim={0pt 10pt 75pt 0pt},clip]{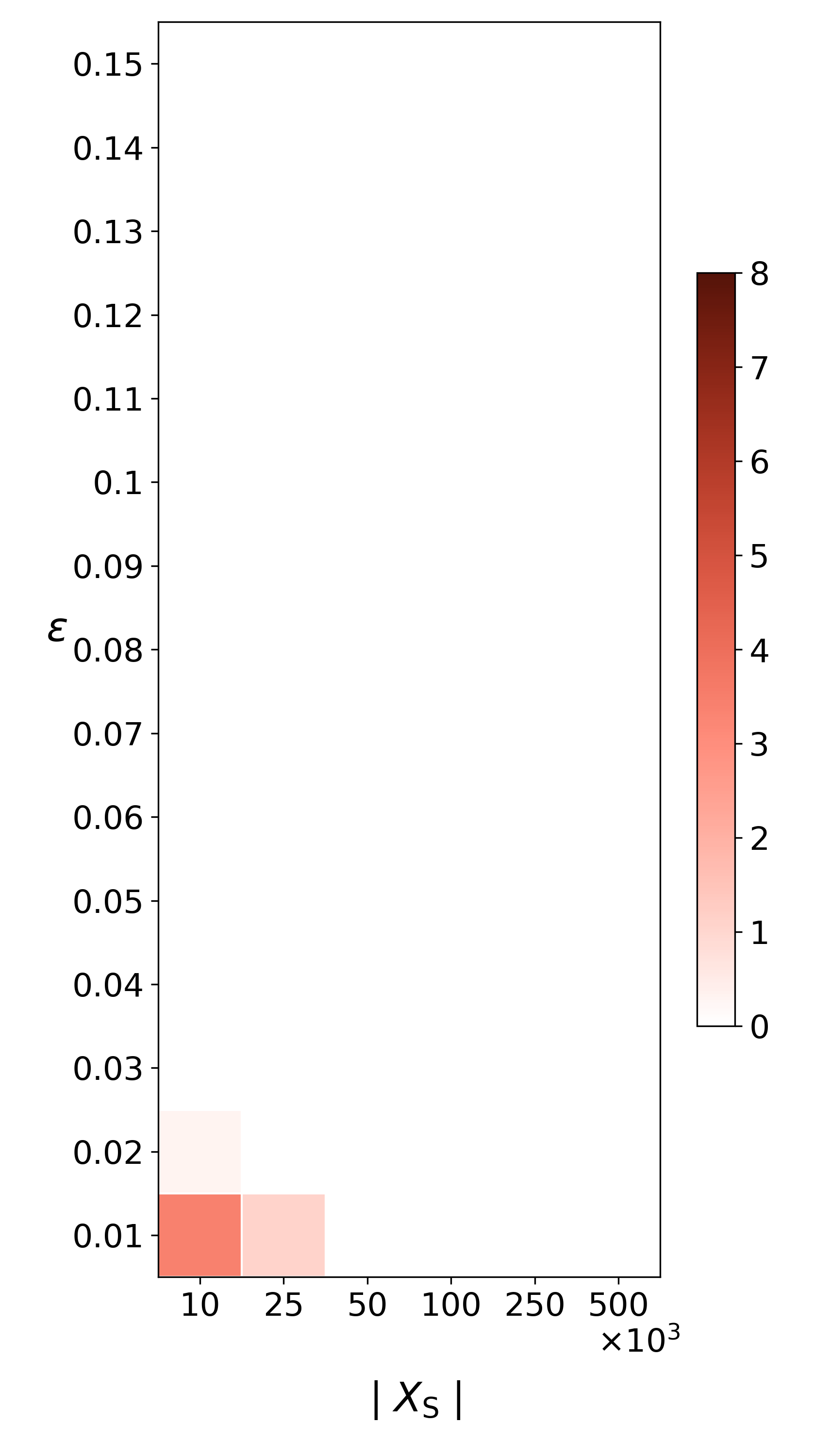}%
    \label{fig:heatmap_c2dtlz2_2}}
    \subfloat[C2-DTLZ2 ($D = 3$)]{\includegraphics[height=227pt,trim={0pt 10pt 75pt 0pt},clip]{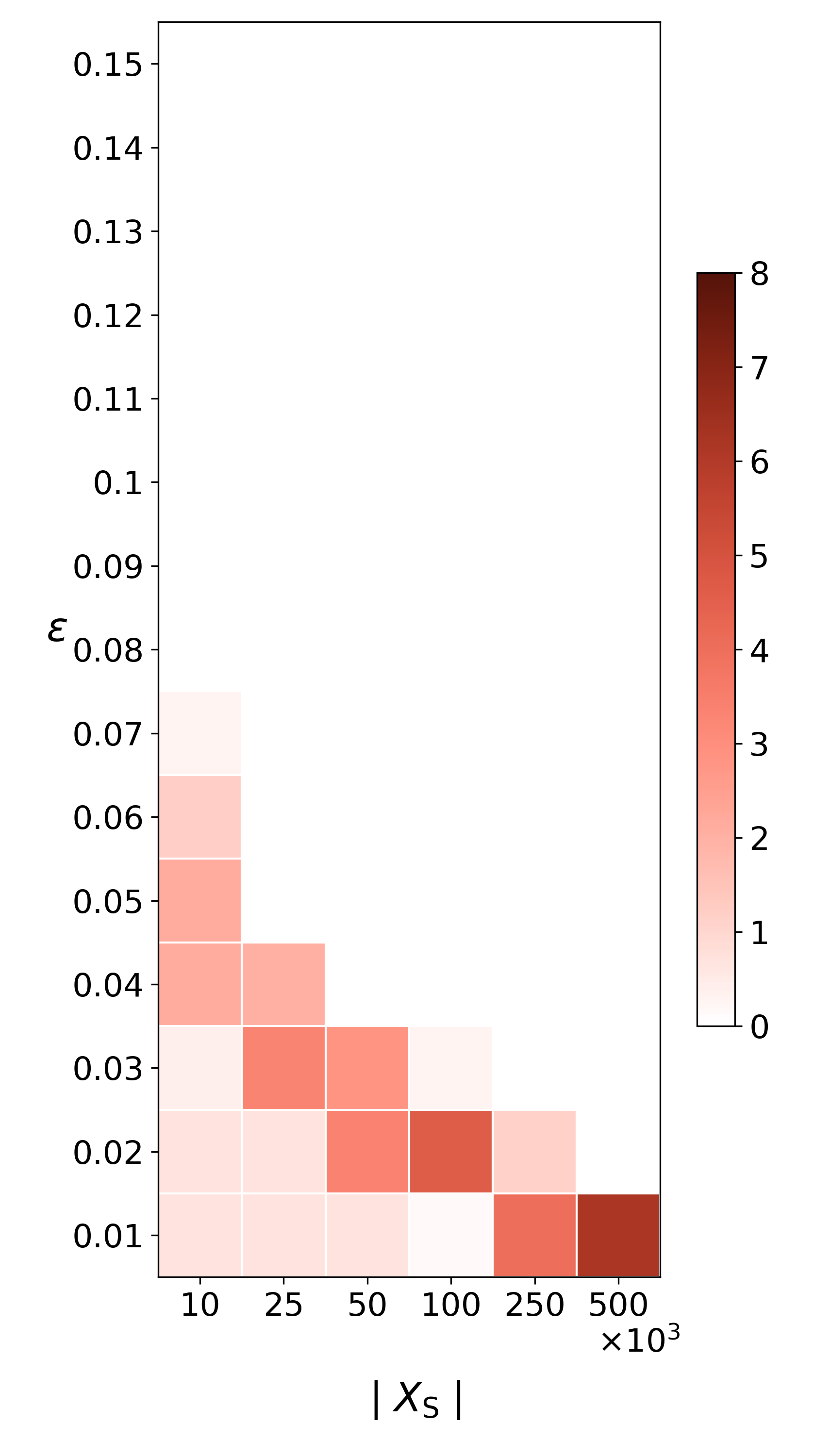}%
    \label{fig:heatmap_c2dtlz2_3}}
    \subfloat[DAS-CMOP1 ($D = 2$)]{\includegraphics[height=227pt,trim={0pt 10pt 75pt 0pt},clip]{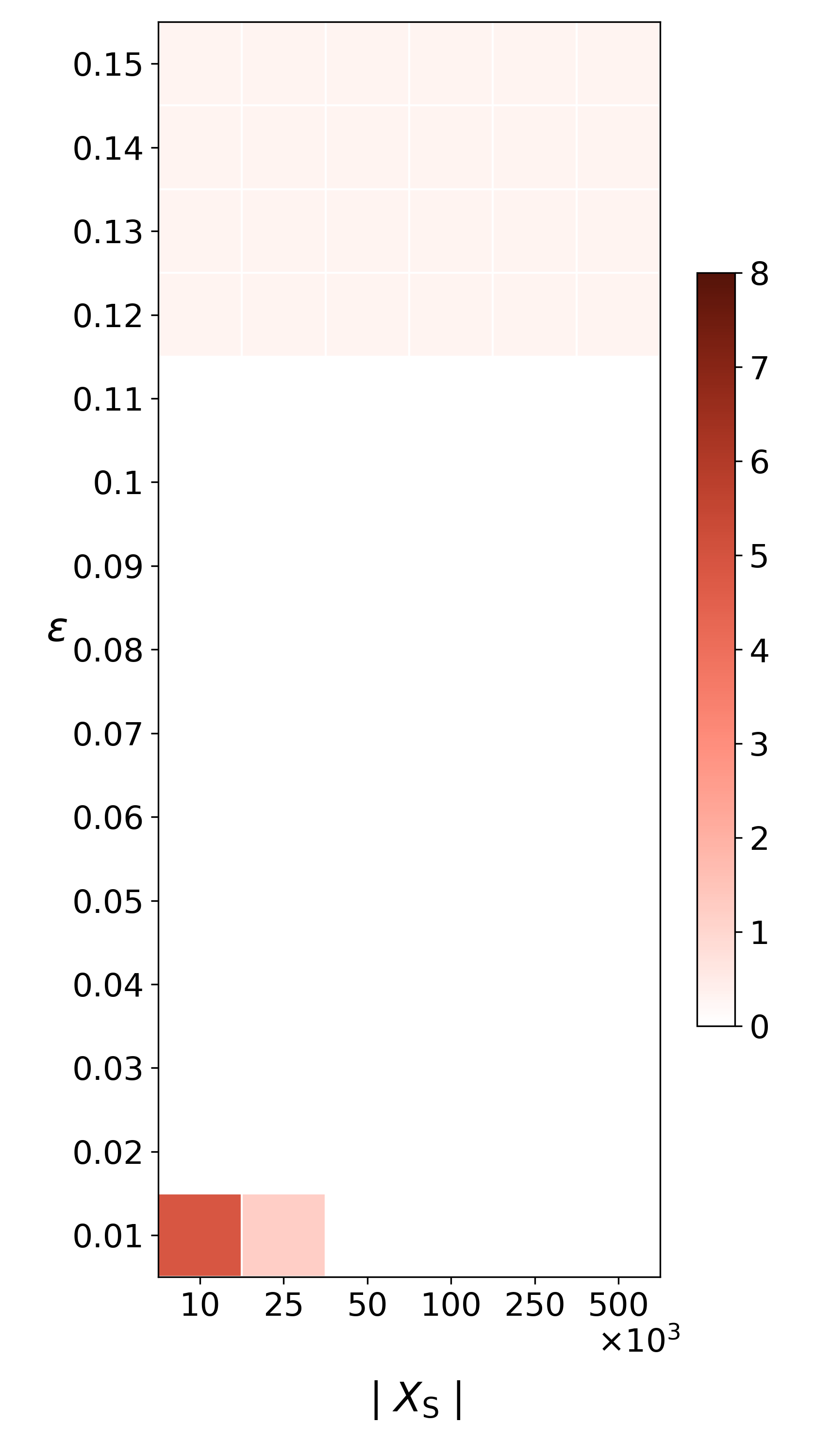}%
    \label{fig:heatmap_dascmop1_2}}
    \subfloat[DAS-CMOP1 ($D = 3$)]{\includegraphics[height=227pt,trim={0pt 10pt 0pt 0pt},clip]{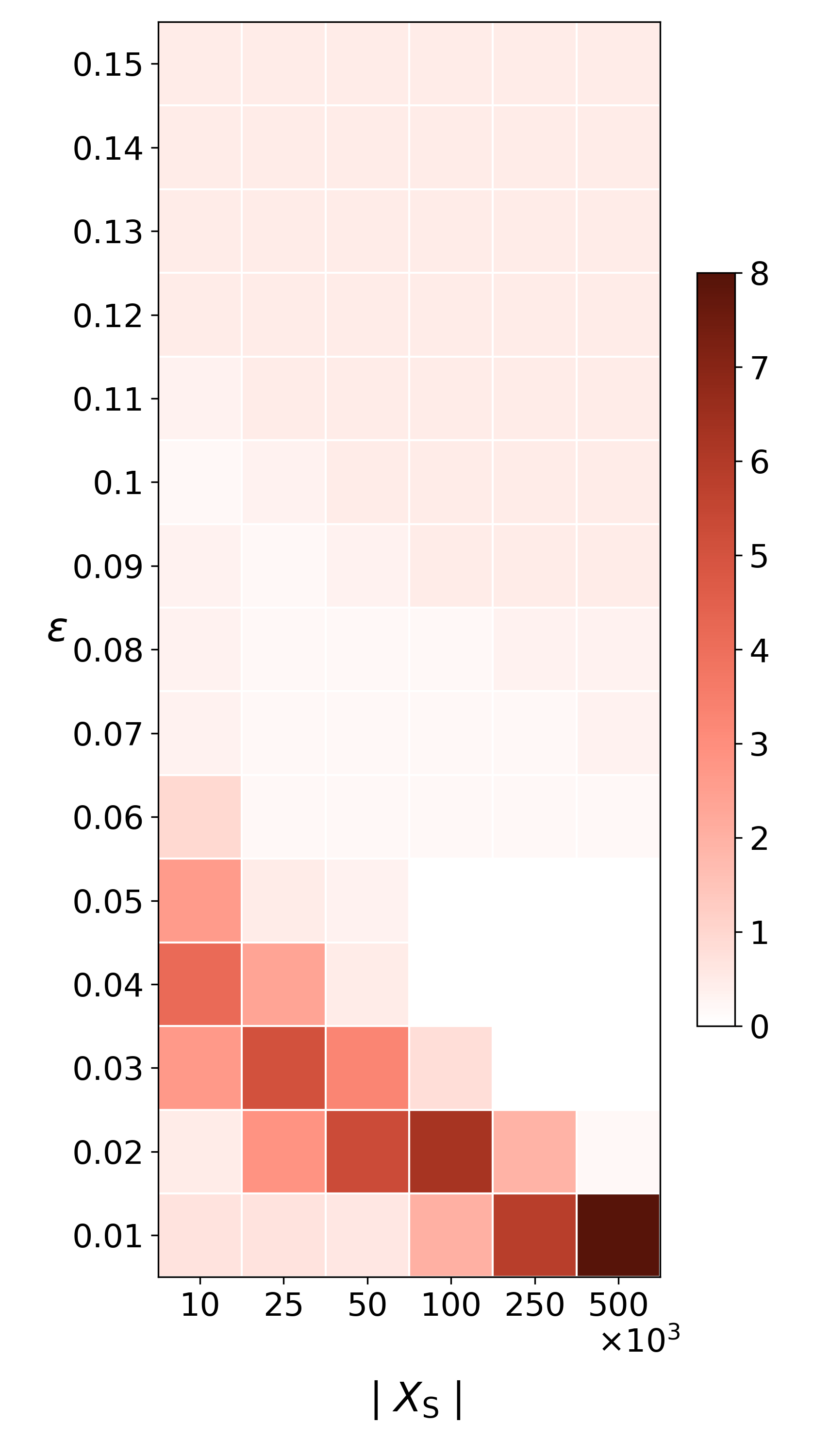}%
    \label{fig:heatmap_dascmop1_3}}
    \caption{Median log differences between the exact number of feasible components and the derived number of clusters for each combination of $\abs{X_\mathrm{S}}$ and $\epsilon$ for the 2-D and 3-D problems C2-DTLZ2 and DAS-CMOP1.}
    \label{fig:heatmap_sensitivity}
\end{figure}

The results are shown in the form of heat maps in Figure~\ref{fig:heatmap_sensitivity}, where the median logarithmic difference between the exact number of feasible components and the derived number of clusters is depicted. The values are normalized with the number of feasible components and logarithmic to obtain better visualization. Smaller values in lighter colors indicate better performance, and vice versa. Specifically, white color indicates a perfect match.

\begin{figure}[!ht]
    \centering
    
    \subfloat[$\varepsilon = 0.02$]{\includegraphics[clip, trim={0, 10pt, 0, 0}, width=0.5\textwidth]{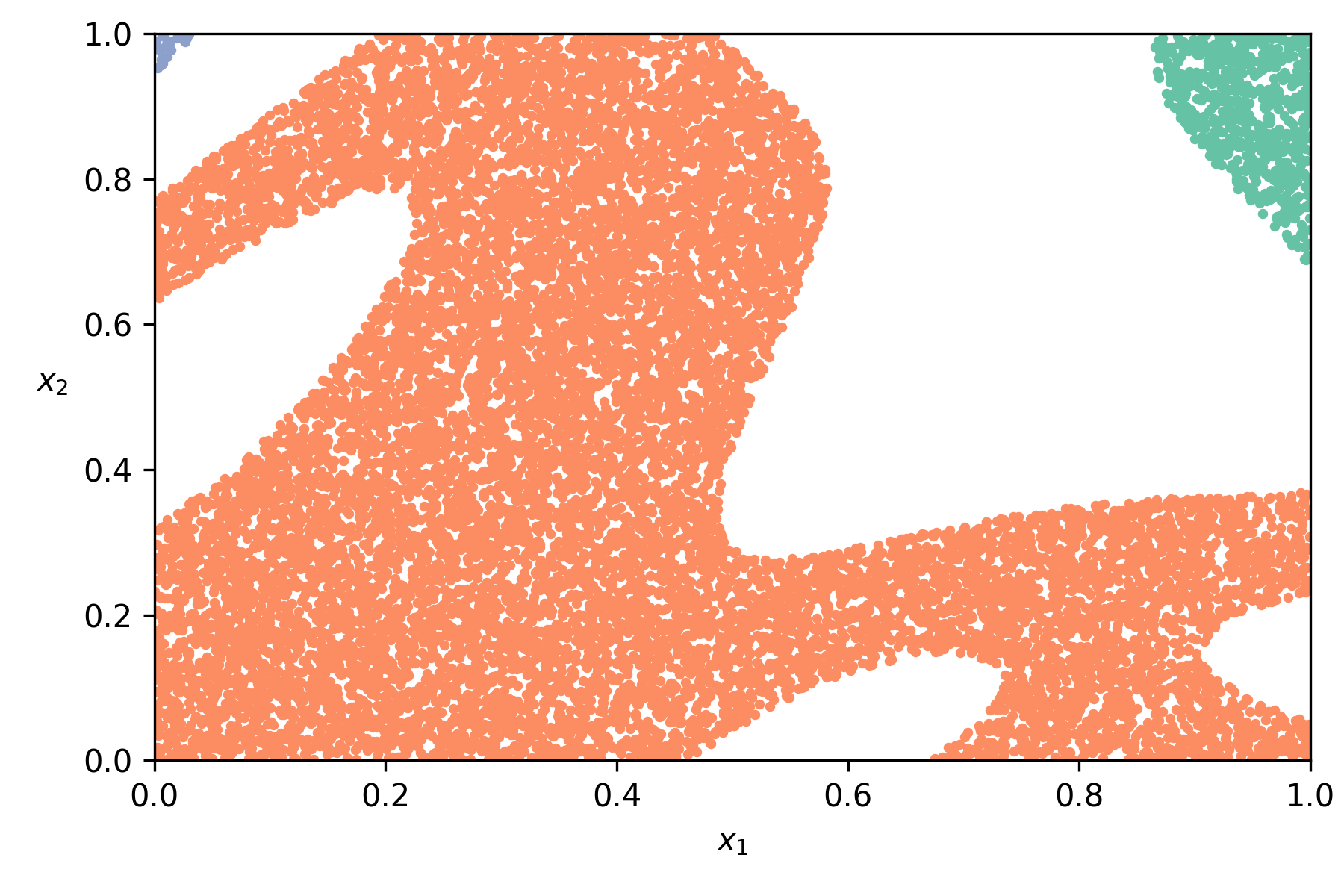}%
    \label{fig:coms_dascmop1_02}}
    \hfil
    \subfloat[$\varepsilon = 0.12$]{\includegraphics[clip, trim={0, 10pt, 0, 0}, width=0.5\textwidth]{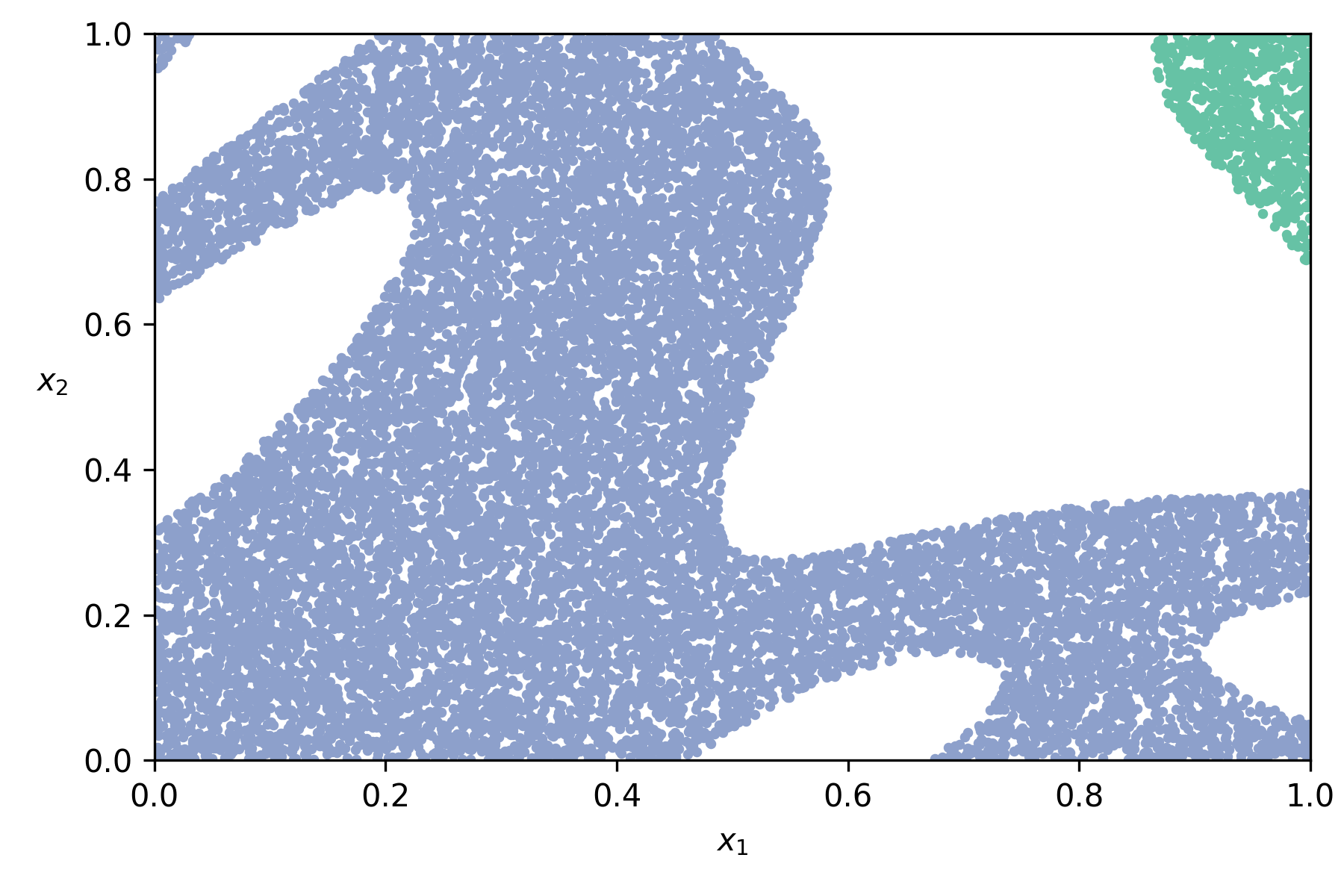}%
    \label{fig:coms_dascmop1_12}}
    
    \caption{Feasible components for DAS-CMOP1 as identified by space-filling design approach for $\varepsilon = 0.02$ (left), $\varepsilon = 0.12$ (right) and $\abs{X_\mathrm{S}} = 25\,000$. Different colors correspond to disjoint feasible components. For $\varepsilon = 0.12$, we can see that the small component in the upper left corner is merged with the large one in the middle.}
    \label{fig:coms}
\end{figure}

As we can see from Figure~\ref{fig:heatmap_c2dtlz2_2}, for C2-DTLZ2 ($D=2$) the exact number of feasible components can be obtained for all $\abs{X_\mathrm{S}} \geq 25\,000$ and all $\varepsilon \geq 0.02$, while for DAS-CMOP1 ($D=2$) already the combination $\abs{X_\mathrm{S}} = 10\,000$ and $\varepsilon = 0.02$ is sufficient to identify the feasible components (Figure~\ref{fig:heatmap_dascmop1_2}). Interestingly, for DAS-CMOP1 ($D=2$) we can see that for $\varepsilon \geq 0.12$ one component is recognized incorrectly. What happens is that the small component in the top left corner is merged with the large one due to the oversized $\varepsilon$ (Figure~\ref{fig:coms}). Nevertheless, a combination of $\abs{X_\mathrm{S}} = 25\,000$ and $\varepsilon = 0.02$ proved to be the most reliable for $D=2$, resulting in a perfect match for all 30 repetitions on C2-DTLZ2 and DAS-CMOP1. Moreover, Figure~\ref{fig:coms} shows that DBSCAN is really effective also when dealing with CMOPs with irregularly-shaped feasible components.

For $D=3$, Figures~\ref{fig:heatmap_c2dtlz2_3}~and~\ref{fig:heatmap_dascmop1_3} clearly show we need larger $\abs{X_\mathrm{S}}$ and $\varepsilon$ to obtain the exact number of components. For both CMOPs a combination of $\abs{X_\mathrm{S}} = 100\,000$ and $\varepsilon = 0.04$ proved to be the most effective, correctly matching all the components in all 30 repetitions. Additionally, we can see that for DAS-CMOP1 two or more components are incorrectly merged for $\varepsilon \geq 0.06$. An interesting observation is that the technique seems to work better for smaller initial sample sizes when using $\varepsilon \leq 0.02$. The reason behind this behavior is that DBSCAN finds no cluster at all in such cases and the number of identified feasible components is thus zero.

In conclusion, for feasible components that are spaced apart, the space-filling design and similarly adaptive walk approaches are highly reliable and insensitive to small changes in parameter values. Furthermore, the performance is not affected by the shape of the feasible components. On the other hand, for problems with feasible components that are placed close together, the techniques are more sensitive to the selection of parameter values.

Finally, a discussion on the parameter sensitivity for information content and random walks can be found in~\cite{Munoz2015b, Malan2015}. For this reason, we just briefly summarize the key findings. In~\cite{Munoz2015b}, a comprehensive sensitivity analysis was performed to measure the robustness of the information content approach. Among others, the authors addressed the effect of varying the size of $X_\mathrm{I}$. The experimental results showed that already a sample size of $100D$ proved to be robust for deriving information content features. Similarly, in~\cite{Malan2015}, the authors demonstrated that the random walk approach was reliable for deriving ratio of feasible boundary crossings. This was shown on a wide variety of violation landscapes by obtaining small standard deviations while repeating the experiments 30 times.

\subsection{A note on scalability}

For problems with up to five variables the derived ELA feature values closely coincided with actual values. In contrast, it was virtually impossible to investigate high-dimensional CMOPs since the size of the initial sample used by the space-filling design techniques grows exponentially. Indeed, a more detailed experimental analysis showed that to obtain meaningful results, we would need at least 2\,000\,000 solutions for problems with seven variables and 15\,000\,000 solutions for problems with ten variables. This is not unexpected since it is a well-known consequence of the curse of dimensionality. 

In addition, the feasible components of the artificial test problems often have the shape of an ellipsoid (Figures~\ref{fig:c2dtlz2_vl}~and~\ref{fig:mw6_vl}). As the dimension of the search space increases, a feasible component (ellipsoid) becomes an insignificant volume relative to that of the decision space (rectangular area). For this reason, it is very hard to obtain relevant results using space-filling design techniques for such large-scale problems.

\section{Conclusions} 
\label{sec:conclusions}

In this work, we have presented a comprehensive investigation of CMOPs using landscape analysis. Firstly, we have extended the concepts of fitness and violation landscapes to constrained multiobjective optimization and proposed a mathematical formulation to describe multimodality in the violation landscape. Secondly, four ELA techniques, namely space-filling design, information content, random walks, and adaptive walks, have been adapted for CMOPs. Next, we have used the resulting features to analyze eight artificial test suites and compare them against the RCM suite, which consists of real-world problems based on physical models. In particular, the artificial test suites were assessed with respect to the representativeness of various RCM characteristics. Additionally, by contrasting the test suites against a set of all the considered CMOPs, we have also assessed whether the test suites contain problems with various characteristics. Finally, we have discussed the advantages and limitations of the existing artificial test suites as well as the sensitivity and limitations of our methodology. The experimental results show that the proposed ELA techniques are adequate to characterize constrained multiobjective optimization problem landscapes with up to five decision variables. 

Our findings indicate that the artificial test problems fail to satisfactorily represent some real-world problem characteristics, e.g., strong negative correlation between the objectives and the overall constraint violation. Moreover, the predominant source of complexity in novel artificial test problems is very often violation multimodality. This can lead to biased interpretations and conclusions since many other relevant characteristics are missed. All this clearly suggests that further advances in landscape analysis for CMOPs are required to overcome these drawbacks.

This study is an important step towards enhancing the understanding of test suites developed for constrained multiobjective optimization. It has revealed that all the studied artificial test suites have certain advantages and limitations and showed that no ``perfect'' test suite exists.

However, we are aware that the presented work has two shortcomings. Firstly, the proposed approach to characterize violation multimodality is not sufficiently scalable, and thus only low-dimensional problems were addressed in this study. Secondly, the correlations between the proposed features and multiobjective optimization algorithm performances were not evaluated. Consequently, the effect of characteristics on constraint handling techniques could not be adequately assessed. To overcome these limitations, we will experiment with ELA techniques to characterize large-scale optimization problems. We will also comprehensively measure the algorithm performance while solving test CMOPs. Special attention will be devoted to understanding the behavior of constraint handling techniques subject to various problem characteristics.

\section*{Acknowledgment}

We acknowledge financial support from the Slovenian Research Agency (young researcher program and research core funding no.\ P2-0209). This work is also part of a project that has received funding from the European Union's Horizon 2020 research and innovation program under Grant Agreement no.\ 692286.

\bibliography{vodo21a}

\end{document}